\documentclass{article}

\usepackage[preprint]{neurips_2025}

\newcommand{\gL}{\mathcal{L}}

\usepackage[utf8]{inputenc} 
\usepackage[T1]{fontenc}    
\usepackage{hyperref}       
\usepackage{url}            
\usepackage{booktabs}       
\usepackage{amsfonts}       
\usepackage{nicefrac}       
\usepackage{microtype}      
\usepackage{xcolor}         
\usepackage{caption}
\usepackage{natbib}
\usepackage{enumitem}
\usepackage{array}
\usepackage{amsmath}   
\usepackage{amssymb}   
\usepackage{mathtools}

\usepackage{wrapfig}
\usepackage{bbm}
\def\Real{\mathbb{R}}
\usepackage{color}

\usepackage{amsmath,amssymb}
\usepackage{amsthm,bm}
\usepackage{cases}
\usepackage[normalem]{ulem}
\usepackage{multirow}

\usepackage{caption}
\usepackage{dsfont}

\usepackage{comment}

\usepackage{subfigure}
\usepackage{graphicx}

\usepackage{algorithm}
\usepackage{algorithmic}
\usepackage{float}

\input{mysymbol.sty}

\title{Filtering Learning Histories Enhances \\ In-Context Reinforcement Learning}

\author{%
  Weiqin Chen$^\spadesuit$ \quad Xinjie Zhang$^\heartsuit
$ \quad Dharmashankar Subramanian$^\clubsuit$ \quad Santiago Paternain$^\spadesuit$  \\ \\
  $^\spadesuit$ Rensselaer Polytechnic Institute \\
  $^\heartsuit$ Columbia University \\
  $^\clubsuit$ IBM Research
}

\begin{document}

\maketitle

\begin{abstract}

Transformer models (TMs) have exhibited remarkable in-context reinforcement learning (ICRL) capabilities, allowing them to generalize to and improve in previously unseen environments without re-training or fine-tuning. This is typically accomplished by imitating the complete learning histories of a source RL algorithm over a substantial amount of pretraining environments, which, however, may transfer suboptimal behaviors inherited from the source algorithm/dataset. Therefore, in this work, we address the issue of inheriting suboptimality from the perspective of dataset preprocessing. Motivated by the success of the weighted empirical risk minimization, we propose a simple yet effective approach, learning history filtering (LHF), to enhance ICRL by reweighting and filtering the learning histories based on their improvement and stability characteristics. To the best of our knowledge, LHF is the first approach to avoid source suboptimality by dataset preprocessing, and can be combined with the current state-of-the-art (SOTA) ICRL algorithms. We substantiate the effectiveness of LHF through a series of experiments conducted on the well-known ICRL benchmarks, encompassing both discrete environments and continuous robotic manipulation tasks, with three SOTA ICRL algorithms (AD, DPT, DICP) as the backbones. LHF exhibits robust performance across a variety of suboptimal scenarios, as well as under varying hyperparameters and sampling strategies. Notably, the superior performance of LHF becomes more pronounced in the presence of noisy data, indicating the significance of filtering learning histories.

\end{abstract}

\section{Introduction}
\label{section_introduction}

For many years now, numerous reinforcement learning (RL) methods, with varying degrees of success, have been developed to address a wide variety of decision-making problems, such as strategy games~\citep{mnih2013playing, mnih2015human}, robotics~\citep{levine2016end, duan2016benchmarking}, and recommender systems~\citep{afsar2022reinforcement, lin2023survey}.
However, RL suffers from a persistent challenge of severe sample inefficiency due to its trial-and-error learning nature~\cite{sutton2018reinforcement}. Moreover, standard RL approaches typically require retraining a policy whenever a new environment is encountered~\cite{beck2023survey}. These limitations significantly hinder the practical deployment of RL in real-world scenarios.
Recently, pretrained transformer models (TMs) have exhibited impressive capability of in-context learning~\citep{dong2022survey, li2023transformers, wei2023larger, wies2024learnability}, which allows to infer and understand the new (unseen) tasks provided with the context information (or prompt) and without the need for re-training or fine-tuning TMs.
With the application of TMs to decision-making problems, in-context reinforcement learning (ICRL)~\citep{laskin2022context, lin2023transformers, sinii2023context, zisman2023emergence, lee2024supervised} emerges, wherein the state-action-reward tuples are treated as contextual information.
Current state-of-the-art (SOTA) ICRL algorithms, such as Algorithm Distillation (AD)~\cite{laskin2022context}, employ a source RL algorithm like PPO~\cite{schulman2017proximal} to train across a substantial amount of RL environments and collect the corresponding learning histories. TMs are then used to distill the RL algorithm by imitating these complete learning histories. The pretrained TMs demonstrate promising ICRL performance when evaluated in previously unseen test environments. This is achieved by learning from trial-and-error experiences and improving in context. 
On the other hand, Decision Pretrained Transformer (DPT)~\cite{lee2024supervised} enables ICRL by performing posterior sampling over the underlying Markov Decision Process (MDP). In this framework, TMs are pretrained to infer the target MDP from a given context dataset and to predict the optimal actions corresponding to the inferred MDP. Notably, DPT allows both the context and query state to be derived from the learning histories, while still requiring the prediction of the corresponding optimal actions. Throughout this paper, we focus on this version of DPT that operates directly on learning histories.

\begin{figure}[ht]
  \centering
  \includegraphics[width=\linewidth]{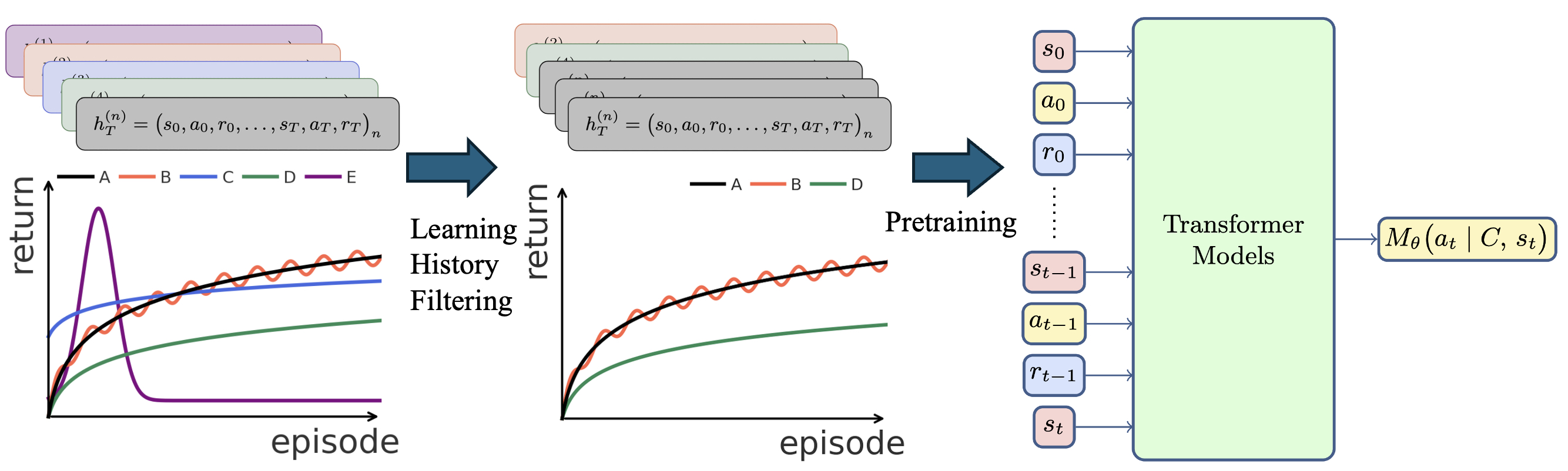}
  \caption{The schematic of learning history filtering (LHF). Current ICRL methods employ a source RL algorithm (e.g., PPO) to collect the learning histories across a substantial amount of environments, resulting in a pretraining dataset composed of multiple learning histories with varying levels of performance (\textit{left}). 
  LHF filters such pretraining dataset and randomly retains each learning history with different probabilities that depend on the improvement and stability characteristics inherent in the learning histories. As a result, high-quality learning histories (A, B, D) are more likely to be retained with varying proportions, while suboptimal ones (C, E) tend to be filtered out (\textit{middle}).
  After filtering learning histories, we follow the standard process for pretraining transformer models (\textit{right}).
    }
  \label{fig_lhf_schematic}
\end{figure}

Despite impressive performance, current SOTA ICRL algorithms often inherit the suboptimal behaviors of the source RL algorithm~\cite{son2025distilling}, as they imitate the entire learning histories. The prior work DICP~\cite{son2025distilling} tackles this challenge by considering an in-context model-based planning framework. Nevertheless, it is significant to emphasize that the suboptimal behaviors embedded within the dataset still adversely affect the performance of ICRL.
Our work thus proposes to address this issue from the perspective of the pretraining dataset.
Motivated by the success of weighted empirical risk minimization (WERM) over standard ERM when guided by appropriate metrics, we filter the pretraining dataset of learning histories by retaining each learning history with a probability depending on its inherent improvement and stability characteristics (refer to Figure~\ref{fig_lhf_schematic}).

\textbf{Main Contributions.} $(i)$ We propose a novel approach of learning history filtering (LHF) to enhance ICRL, which, to the best of our knowledge, is the first method that addresses the issue of inheriting the source suboptimality from the perspective of dataset preprocessing (filtering).
$(ii)$ We substantiate the efficacy of LHF on multiple popular ICRL benchmark environments, including the discrete environments like \textit{Darkroom}-type problems and continuous robotic manipulation tasks, i.e., \textit{Meta-World-ML1}.
Our empirical results demonstrate that LHF consistently outperforms the original baselines without learning history filtering across all backbones algorithms and problems. In certain problems, such as \textit{Reach-Wall} in \textit{Meta-World-ML1}, our LHF approach outperforms the baselines by achieving over a $141\%$ performance enhancement.
$(iii)$ We further validate the robustness of LHF across multiple suboptimal scenarios such as noisy dataset, partial learning histories, and lightweight models, as well as with respect to the hyperparameter variations and different sampling strategies.

\section{Related Work}
\label{section_related_work}

\textbf{Transformer Models for RL.}
TMs~\citep{vaswani2017attention} have been successfully applied to offline RL by their promising capability in sequential modeling. The pioneering works include Decision Transformer~\cite{chen2021decision}, Trajectory Transformer~\cite{janner2021offline}, etc. Specifically, TMs autoregressively model the sequence of actions from the historical offline data conditioned on the sequence of returns in the history.
During the test, the trained model can be queried based on pre-defined target returns, allowing it to generate actions aligned with the target returns.
In addition, Multi-Game Decision Transformer (MGDT)~\citep{lee2022multi} and Gato~\citep{reed2022generalist} have exhibited the success of the autoregressive TMs in learning multi-task policies by fine-tuning or leveraging expert demonstrations in downstream environments. However, they suffer from poor zero-shot generalization and inferior in-context learning capabilities.


\textbf{In-Context Reinforcement Learning.}
The pioneering contributions in the field of ICRL include AD~\cite{laskin2022context} and DPT~\cite{lee2024supervised}, where the former imitates the complete learning histories of a source RL algorithm over a substantial amount of pretraining environments to distill the policy improvement operator, and the latter pretrains a TM to infer the target MDP from the surrounding context (can be derived from the learning history) and to take actions according to the optimal policy for the inferred target MDP.
%
%
%
%
Although prior work~\cite{son2025distilling} has explored in-context model-based planning to address source suboptimality, it falls short of fully resolving the issue. On the other hand, recent studies~\cite{zisman2024emergence, dong2024context, chen2025random} consistently demonstrate that the performance of ICRL remains highly sensitive to the pretraining dataset.
Therefore, our work aims to address the issue of inheriting the source suboptimality from the perspective of dataset preprocessing.

\textbf{Weighted Empirical Risk Minimization.}
It is worth noting that ICRL follows a supervised pretraining mechanism~\cite{chen2025random}, which essentially undergoes an ERM process~\cite{bartlett2005local, massart2006risk}.
ERM identifies an optimal hypothesis from a hypothesis class that minimizes the empirical risk given a set of (input, label) samples. \cite{zhang2025reweighting} presents a WERM schema that exhibits provably improved excess risk bounds on ``high confidence'' regions than that of standard ERM. These ``high confidence'' regions could be large-margin regions in classification tasks and low-variance regions in heteroscedastic bounded regression problems~\cite{zhang2025reweighting}. Motivated by the superior performance of WERM over the standard ERM, we propose to preprocess (filter) the ICRL pretraining dataset by emulating a WERM schema, combined with crucial aspects in the ICRL like the improvement and stability of the learning histories.


\section{In-Context Reinforcement Learning}
\label{section_preliminary}


\textbf{RL Preliminaries.}
RL is a data-driven solution to MDPs~\citep{sutton2018reinforcement}. 
An MDP can be represented by a tuple $\tau = (\mbS, \mbA, R, P, \rho)$, where $\mbS$ and $\mbA$ denote state and action spaces, $R: \mbS \times \mbA \to \Real$ denotes the reward function that evaluates the quality of the action, $P: \mbS \times \mbA \times \mbS \to [0, 1]$ denotes the transition probability that describes the dynamics of the system, and $\rho: \mbS \to [0, 1]$ denotes the initial state distribution.  
A policy $\pi$ defines a mapping from the states to the probability distributions over the actions, providing a strategy that guides the agent in the decision-making.
The agent interacts with the environment following the policy $\pi$ and the transition dynamics of the system, and then generates a learning history $(s_0, a_0, r_0, s_1, a_1, r_1, \cdots)$.
The performance measure $J (\pi)$ is defined by the expected discounted cumulative reward under the policy $\pi$
\begin{align}
    J (\pi) = \mbE_{s_0 \sim \rho, a_t \sim \pi(\cdot | s_t), s_{t+1} \sim P(\cdot|s_t, a_t)} \left[ \sum_{t=0}^\infty \gamma^t r_t \right].
\end{align}
The goal of RL is to identify an optimal policy $\pi^\star$ that maximizes $J (\pi)$.
It is crucial to recognize that $\pi^\star$ often varies across different MDPs (environments). Accordingly, the optimal policy for standard RL must be re-learned each time a new environment is encountered.
To overcome this limitation, ICRL pretrains a TM on a wide variety of environments, and then deploys it in the \textit{unseen} test environments without updating the parameters of the TM, i.e., zero-shot generalization~\citep{sohn2018hierarchical, mazoure2022improving, zisselman2023explore, kirk2023survey}.


\textbf{Supervised Pretraining of ICRL.}
%
%
%
Consider two distributions over the environments $\ccalT_{\text{pretrain}}$ and $\ccalT_{\text{test}}$ for pretraining and test. Each environment, along with its corresponding MDP $\tau$, can be regarded as an instance drawn from the environment distributions, where each environment may exhibit distinct reward functions and transition dynamics. Given an environment $\tau$, a context $\ccalC = \{s_i, a_i, r_i\}_{i \in [n']}$ refers to a collection of interactions between the agent and the environment $\tau$, sampled from a context distribution $\ccalD_{\text{pretrain}} (\cdot \mid \tau)$, i.e., $\ccalC \sim \ccalD_{\text{pretrain}} (\cdot \mid \tau)$. Notably, $\ccalD_{\text{pretrain}} (\cdot \mid \tau)$ contains the contextual information regarding the environment $\tau$.
We next consider a query state distribution $\ccalD_q^\tau$ and a label policy that maps the query state $s_q$ to the distribution of the action label $a_l$, i.e., $\pi_{l}: \mbS \to \Delta_{a_l} (\mbA)$. 
The joint distribution over the environment $\tau$, context $\ccalC$, query state $s_{q}$, and action label $a_{l}$ is given by
\begin{align}\label{eqn_pretrain_distribution}
    \ccalP_{\text{pretrain}} (\tau, \ccalC, s_{q}, a_{l}) = \ccalT_{\text{pretrain}} (\tau) \cdot \ccalD_{\text{pretrain}} (\ccalC | \tau) \cdot \ccalD_q^\tau \cdot \pi_{l}(a_{l} | s_{q}).
\end{align}
The supervised pretraining schema of ICRL is embodied in the process where a TM parameterized by $\theta$ (denoted as $M_\theta: \mbC \times \mbS \to \Delta(\mbA)$ is pretrained to predict the action label $a_l$ given the context $\ccalC$ and the query state $s_q$. To this end, current ICRL methods~\citep{laskin2022context, lee2024supervised, chen2025random, son2025distilling} consider a common objective
\begin{align}\label{eqn_icrl_obj_unweighted}
    \theta^\star = \argmin_\theta \mbE_{\ccalP_{\text{pretrain}}} \left[ l \left( M_\theta(\cdot \mid \ccalC, s_{q}), a_{l}\right) \right],
\end{align}
where $l(\cdot, \cdot)$ represents the loss function, e.g., $l \left( M_\theta(\cdot \mid \ccalC, s_{q}), a_{l}\right) = -
\log M_\theta(a_l \mid \ccalC, s_{q})$. 
%

It is crucial to highlight that, in the context of this work, $\ccalP_{\text{pretrain}}$ describes the distribution of learning histories. While the general problem of ICRL assumes a generic distribution, in this work, the context, query state and action label are obtained from the learning histories of an RL algorithm, e.g., PPO.

%


\section{Learning History Filtering}
\label{section_LHF}

This section presents our dataset preprocessing approach, learning history filtering (LHF; summarized in Algorithm~\ref{alg_lhf}), which is inspired by the success of WERM~\cite{zhang2025reweighting} and the fact that ICRL adheres to a supervised pretraining paradigm. Specifically, WERM demonstrates that reweighting the training objective based on appropriate metrics can lead to provable performance enhancement.
In the remainder of this section, we start by presenting a weighted learning history sampling mechanism for ICRL that emulates the WERM schema. Following that, we formally define the metrics used in the weighted sampling that play important roles in the pretraining of ICRL (supported by our empirical evidence in Section~\ref{section_experiment}). Lastly, we describe specific sampling strategies based on these metrics that are equivalent to weighting the learning histories.

\textbf{Weighted Sampling for ICRL.}
%
%
%
%
%
%
WERM~\cite{zhang2025reweighting} leverages a problem-dependent weighted structure to improve upon ERM. Concretely, an input-dependent weight function $w(\cdot)$ is employed to re-weight the ERM objective. In the case of ICRL, the new weighted objective function based on \eqref{eqn_icrl_obj_unweighted} is given by
\begin{align}\label{eqn_icrl_obj_weighted_train}
    \theta_w^\star = \argmin_\theta \mbE_{\ccalP_{\text{pretrain}}} \left[ w(\tau, \mathcal{C}, s_q, a_l) \cdot l \left( M_\theta(\cdot \mid \ccalC, s_{q}), a_{l}\right) \right],
\end{align}
where the weight $w$ relies on the environment $\tau$, context $\ccalC$, query state $s_q$, and action label $a_l$, and is essentially determined by the learning history.
To emulate the WERM schema during the dataset preprocessing, we adopt a random sampling strategy guided by the learning history. In particular, we define $\bar{w}$ to be the random variables taking the values $0$ or $1$ according to a distribution that we denote by $\mathcal{P}_{\bar{w}}(\tau,\mathcal{C},s_q,a_l)$ (see e.g., \eqref{eqn_linear_weight_func}). Subsequently, by defining a new learning history distribution $\ccalP_{\text{pretrain}}^w = \mathcal{P}_{\bar{w}}(\tau, \mathcal{C}, s_q, a_l) \cdot \ccalP_{\text{pretrain}} (\tau, \ccalC, s_{q}, a_{l})$, we adopt the following objective
\begin{align}\label{eqn_icrl_obj_weighted_data}
    \theta_w^\star = \argmin_\theta \mbE_{\ccalP_{\text{pretrain}}^w} \left[ l \left( M_\theta(\cdot \mid \ccalC, s_{q}), a_{l}\right) \right].
\end{align}
Notice that for any set of weights in \eqref{eqn_icrl_obj_weighted_train} between $0$ and $1$, it is always possible to define a probability distribution $\mathcal{P}_{\bar{w}}(\tau,\mathcal{C},s_q,a_l)$ such that \eqref{eqn_icrl_obj_weighted_data} becomes equivalent to  \eqref{eqn_icrl_obj_weighted_train}.

\textbf{Improvement and Stability of Learning Histories.}
%
%
%
%
The weight in the WERM is crafted to reflect key aspects of the training process, such as the improvement and stability characteristics inherent in learning trajectories as exemplified by ICRL~\cite{laskin2022context, zisman2024emergence}.
To formalize the improvement and stability, we define a learning history as the collection of state-action-reward tuples $(s, a, r)$ generated during a single run of a source RL algorithm (e.g., PPO) within a single environment. Since each environment may yield multiple learning histories, we denote by $\ccalD_i^l$ the $l$-th learning history in the $i$-th environment.
Then, we define the improvement of a learning history $\ccalD_i^l$ with respect to its episodic returns 
\begin{align}
    \texttt{Improvement}(\ccalD_i^l) = \frac{\bar{R}(\ccalD_i^l) + R_G(\ccalD_i^l)}{2 R_\text{max}^i}, 
\end{align}
where $R_\text{max}^i$ denotes the maximum episodic return available in the $i$-th environment, $\bar{R}(\ccalD_i^l)$ represents the mean of episodic returns in the learning history $\ccalD_i^l$, and $R_G(\ccalD_i^l)$ denotes the difference (gap) between the maximal and minimal episodic returns in the learning history $\ccalD_i^l$. Note that the \texttt{Improvement} metric takes a value in $[0, 1]$.

To quantify the stability of a learning history, we consider the sequence of episodic returns within $\ccalD_i^l$ and compute the difference between each return and its immediate successor. We then extract the negative differences, indicating performance degradations, and compute their mean. This measure is denoted by $\bar{R}_D(\ccalD_i^l)$. Subsequently, we define the stability of the learning history $\ccalD_i^l$ by
\begin{align}
    \texttt{Stability}(\ccalD_i^l) = 1 + \frac{\bar{R}_D(\ccalD_i^l)}{R_\text{max}^i},
\end{align}
where the $\texttt{Stability}$ metric takes a value in the range $[0, 1]$ as well.
Having formalized the improvement and stability, we integrate them into a unified metric
\begin{align}\label{eqn_metric}
    U(\ccalD_i^l) = \texttt{Improvement}(\ccalD_i^l) + \lambda \cdot \texttt{Stability}(\ccalD_i^l),
\end{align}
where $\lambda$ is a hyperparameter that trades-off the improvement and stability. Indeed, for large values of $\lambda$ the unified metric $U(\ccalD_i^l)$ will prioritize the stability in the learning history, whereas for small values of $\lambda$ the metric will focus on the improvement. Section~\ref{section_sensitivity} demonstrates the robust performance of our LHF approach with respect to various choices of $\lambda$. As depicted in Figure~\ref{fig_lhf_schematic}, $U(\ccalD_i^l)$ encapsulates important characteristics in the learning history that play crucial roles in the pretraining of TMs.

\textbf{Sampling Strategy.}
Having defined the unified metric $U(\ccalD_i^l)$, we are now in the stage of introducing the sampling strategy that allows us to emulate the WERM scheme during the dataset preprocessing.
\begin{algorithm}
\caption{Learning History Filtering (LHF)}         
\label{alg_lhf}
\begin{algorithmic}[1]

        \STATE \textbf{Require:} Pretraining dataset $\{\ccalD_i^l\}$ with $i \in [N_i], l \in [N_l]$, empty LHF dataset $\ccalD_\text{LHF}$

        \FOR{$i$ in $[N_i]$}

            \STATE Let $\ccalD_i' = \emptyset$
            \WHILE{$|\ccalD_i'| < |\ccalD_i|$}
                    
                \FOR{$l$ in $[N_l]$}

                    \STATE Compute the unified metric $U(\ccalD_i^l)$ by \eqref{eqn_metric}
                    \STATE Compute the weighted probability $\mathcal{P}_{\bar{w}}(U(\ccalD_i^l))$ for the learning history $\ccalD_i^l$ by \eqref{eqn_linear_weight_func}
                    \STATE Sample a uniform random variable $v \sim \ccalU[0, 1]$
                    \STATE Add the learning history $\ccalD_i^l$ to $\ccalD_i'$ \textbf{if} $v \leq \mathcal{P}_{\bar{w}}(U(\ccalD_i^l))$
                    \IF{$|\ccalD_i'| = |\ccalD_i|$}
                    \STATE break
                    \ENDIF
                \ENDFOR
            \ENDWHILE
 
            \STATE $\ccalD_\text{LHF} \leftarrow \ccalD_\text{LHF} \cup \ccalD_i'$
        \ENDFOR
        
        \STATE \textbf{Return} $\ccalD_\text{LHF}$
\end{algorithmic}
\end{algorithm}

Given a static pretraining dataset $\{\ccalD_i^l\}$, where $i \in [N_i]$ indexes environments and $l \in [N_l]$ indexes learning histories within each environment, we construct an empty dataset $\ccalD_{\text{LHF}}$ for filtering the learning history. For each learning history $\ccalD_i^l$, we define a weighted sampling probability that depends linearly on its unified metric $U(\ccalD_i^l)$
\begin{align}\label{eqn_linear_weight_func}
    \mathcal{P}_{\bar{w}}(U(\ccalD_i^l)) = \frac{U(\ccalD_i^l) - \min_{l \in [N_l]} U(\ccalD_i^l)}{\max_{l \in [N_l]} U(\ccalD_i^l) - \min_{l \in [N_l]} U(\ccalD_i^l)}.
\end{align}
Guided by \eqref{eqn_icrl_obj_weighted_data}, we in turn randomly select the learning histories in $\{\ccalD_i^l\}$ with the corresponding weighted probability $\mathcal{P}_{\bar{w}}(U(\ccalD_i^l))$ for each learning history in each environment, and add it to our LHF dataset $\ccalD_\text{LHF}$ until its size matches that of $\{\ccalD_i^l\}$. The procedure is detailed in Algorithm~\ref{alg_lhf}.
It is important to note that the linear sampling strategy \eqref{eqn_linear_weight_func} is not the only choice for our LHF approach. Other sampling strategies, such as Softmax, can also be employed. Section~\ref{section_sensitivity} demonstrates the robustness of LHF combined with Softmax sampling function with varying temperature parameters.

After preprocessing the dataset by LHF, we follow the standard processes for pretraining and testing TMs as in the ICRL literature~\cite{laskin2022context, lee2024supervised, chen2025random}, which are outlined in Algorithm~\ref{alg_whole} in Appendix~\ref{append_complete_pseudocode}.

\section{Experiments}
\label{section_experiment}

We substantiate the efficacy of our LHF approach across a diverse set of environments, which are commonly considered in ICRL literature~\citep{laskin2022context, lee2024supervised, son2025distilling, chen2025random}.
These environments include discrete settings such as \textit{Darkroom, Darkroom-Permuted, Darkroom-Large, Dark Key-to-Door} and continuous robotic manipulation tasks from the \textit{Meta-World-ML1} benchmark like \textit{Reach, Reach-Wall, Button-Press, Basketball, Door-Unlock, Push, Soccer, Hand-Insert}.
All these problems are challenging to solve in-context, as the test environments differ from the pretraining environments, while the parameters of the TM remain frozen during the test. The environmental setup is detailed in Appendix~\ref{append_env_setup}.

\subsection{Collecting and Filtering Learning Histories}
\label{subsec_collect_filter_hist}

Following previous ICRL works~\cite{laskin2022context, son2025distilling}, we consider PPO as the source RL algorithm to collect learning histories in the \textit{Darkroom}-type and \textit{Meta-World-ML1} problems. As introduced in Appendix~\ref{append_env_setup}, each problem includes multiple distinct environments depending on e.g., the goal locations. For each environment, we employ $100$ PPO agents to collect $100$ learning histories with each comprising $1000$ transitions for \textit{Darkroom}-type and $10,000$ transitions for \textit{Meta-World-ML1}. This yields a total of $100,000$ transitions per \textit{Darkroom}-type environment and $1,000,000$ transitions per \textit{Meta-World-ML1} environment. We provide the detailed procedure of collecting learning histories in Appendix~\ref{append_collect_history}.
Having collected the pretraining dataset of learning histories, we next filter the dataset by LHF, which is detailed in Section~\ref{section_LHF} and summarized in Algorithm~\ref{alg_lhf}.

\subsection{Backbone ICRL Algorithms}

Since our LHF approach exclusively targets the dataset preprocessing, it can be seamlessly integrated with various backbone algorithms to enable ICRL.
In this work, we adopt three SOTA ICRL algorithms (AD, DICP, DPT) as the backbones, each employing distinct strategies to learn from the pretraining dataset of learning histories. More details of the backbone ICRL algorithms are presented in Appendix~\ref{append_backboneICRLalgo}.
Same as in the DICP paper~\cite{son2025distilling}, we assess AD and DICP across all environments introduced in Appendix~\ref{append_env_setup} and evaluate DPT only within the four \textit{Darkroom}-type environments, as DPT relies on the optimal action labels that are typically unavailable in more general environments such as \textit{Meta-World-ML1}.
In addition, since all backbone ICRL algorithms are transformer-based, we consider the same transformer architecture (TinyLlama~\cite{zhang2024tinyllama}) across all experiments to ensure a fair comparison. The transformer hyperparameters, such as the number of attention layers, the number of attention heads, the embedding dimension etc, are detailed in Appendix~\ref{append_transformer_hyperparameter}.

\subsection{Numerical Results}
\label{exp_numerical_result}

\begin{wraptable}{r}{0.5\textwidth}
    \centering
    \caption{Relative enhancement (\%) of LHF over baselines. Backbone algorithms: AD, DICP, DPT.}
    \label{tab_darkrooms_OriginaData_FullHistory}
    \begin{tabular}{cccc}    
        \toprule
        Task & AD & DICP & DPT \\
        \midrule
        \textit{DarkRoom} & \textbf{25.1} & 15.2 & 3.5 \\
        \textit{Darkroom-Permuted} & 5.0 & \textbf{9.2} & 2.6 \\
        \textit{Darkroom-Large} & 3.2 & 9.3 & \textbf{26.1} \\
        \textit{Dark Key-to-Door} & 1.8 & 2.5 & \textbf{15.5} \\
        \midrule
        Average & 8.8 & 9.1 & \textbf{11.9} \\
        \bottomrule
    \end{tabular}
    \vspace{-5pt}
\end{wraptable}

We first exhibit the enhanced performance of our LHF approach by empirical evidence across the three SOTA backbone algorithms and the four \textit{Darkroom}-type environments. Then we move on to the experiments in suboptimal scenarios in terms of the noisy dataset, lightweight model, and partial learning histories, which consistently substantiate the robustness of LHF. Notably, the superiority of LHF becomes even more pronounced in the noisy scenario and across all suboptimal scenarios with AD as the backbone. 
To assess the overall performance of our LHF approach, we adopt the linear sampling strategy and fix the stability coefficient at $\lambda = 1$ across all experiments. That being said, we also examine the robustness of LHF in terms of the varying stability coefficient $\lambda$ and by exploring an alternative Softmax sampling strategy with a set of temperature parameters.
To quantify the relative enhancement of our LHF approach over the original backbone algorithms (baselines) in terms of the speed and final performance, we define the relative enhancement $E$ as $E = (\bar{R}(\text{LHF}) - \bar{R}(\text{baseline})) / \bar{R}(\text{baseline})$ where $\bar{R}(\cdot)$ denotes the mean of episodic returns during the test.

\begin{figure}[ht]
\centering 
\setcounter{subfigure}{0}
\subfigure[Darkroom]
{
	\begin{minipage}{0.25\linewidth}
	\centering 
	\includegraphics[width=1.0\columnwidth]{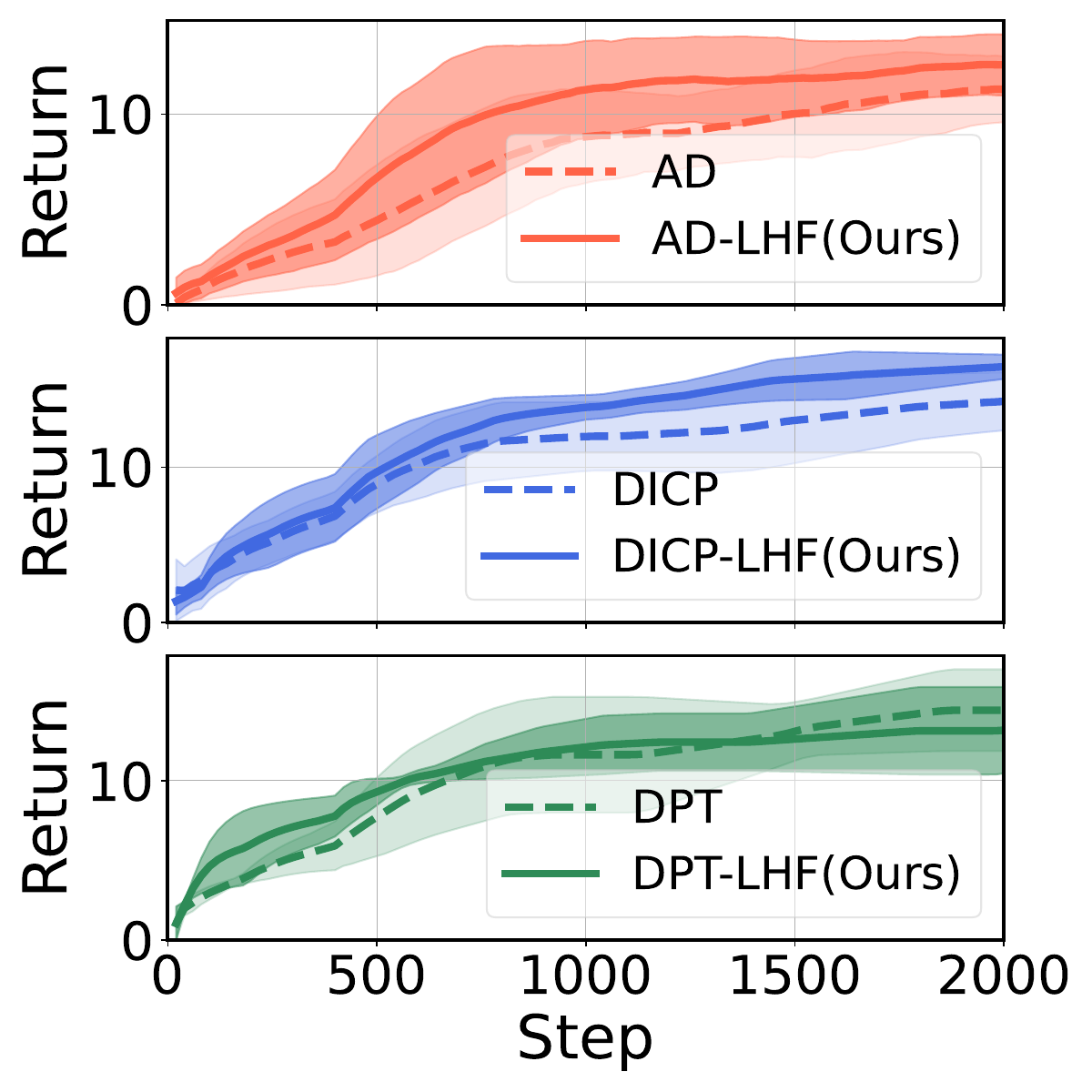}
	\end{minipage}
}
\!\!\!\!\!\!\!
\subfigure[Darkroom-Permuted]
{
	\begin{minipage}{0.25\linewidth}
	\centering 
	\includegraphics[width=1.0\columnwidth]{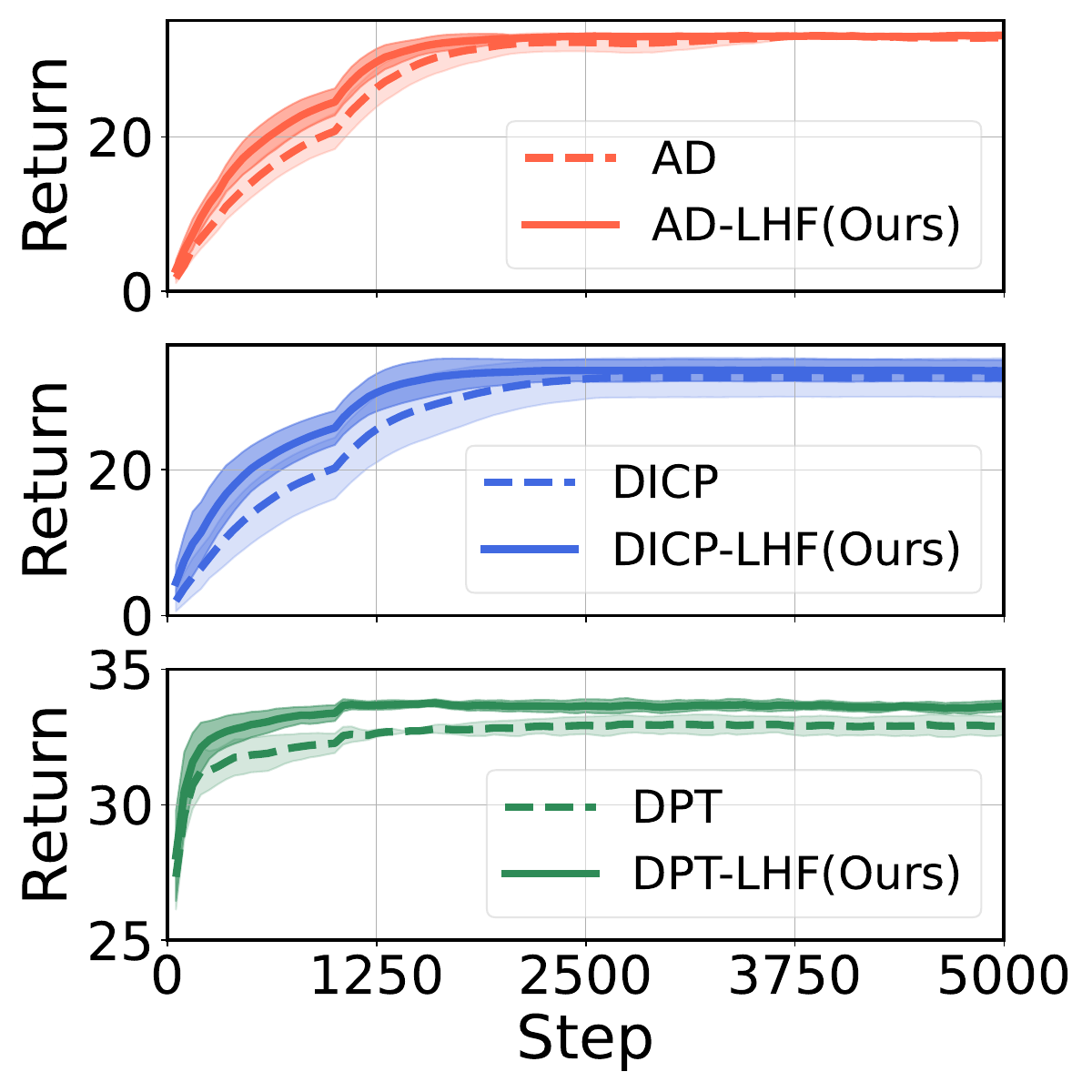}
	\end{minipage}
}
\!\!\!\!\!\!\!
\subfigure[Darkroom-Large]
{
	\begin{minipage}{0.25\linewidth}
	\centering 
	\includegraphics[width=1.0\columnwidth]{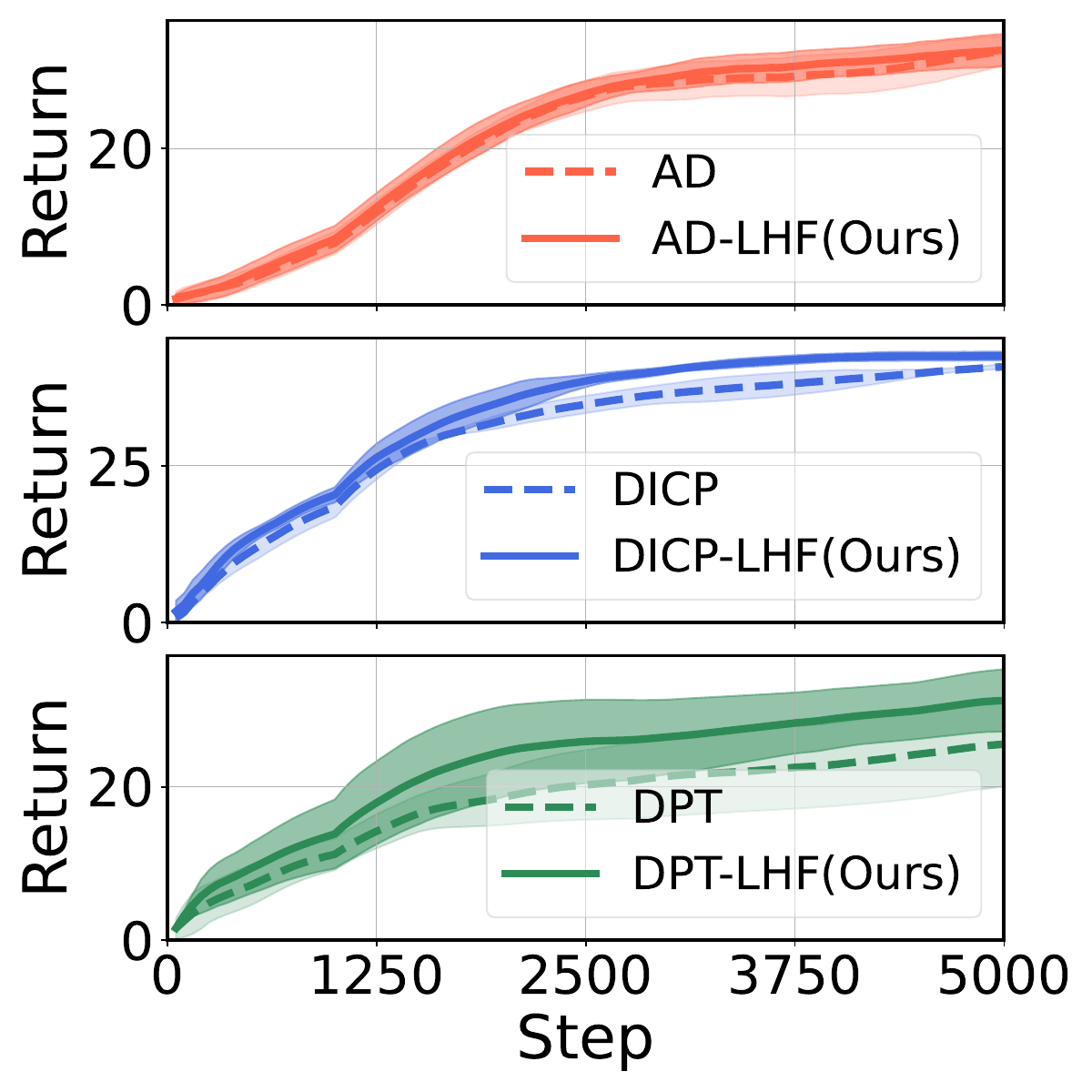}
	\end{minipage}
}
\!\!\!\!\!\!\!
\subfigure[Dark Key-to-Door]
{
	\begin{minipage}{0.25\linewidth}
	\centering 
	\includegraphics[width=1.0\columnwidth]{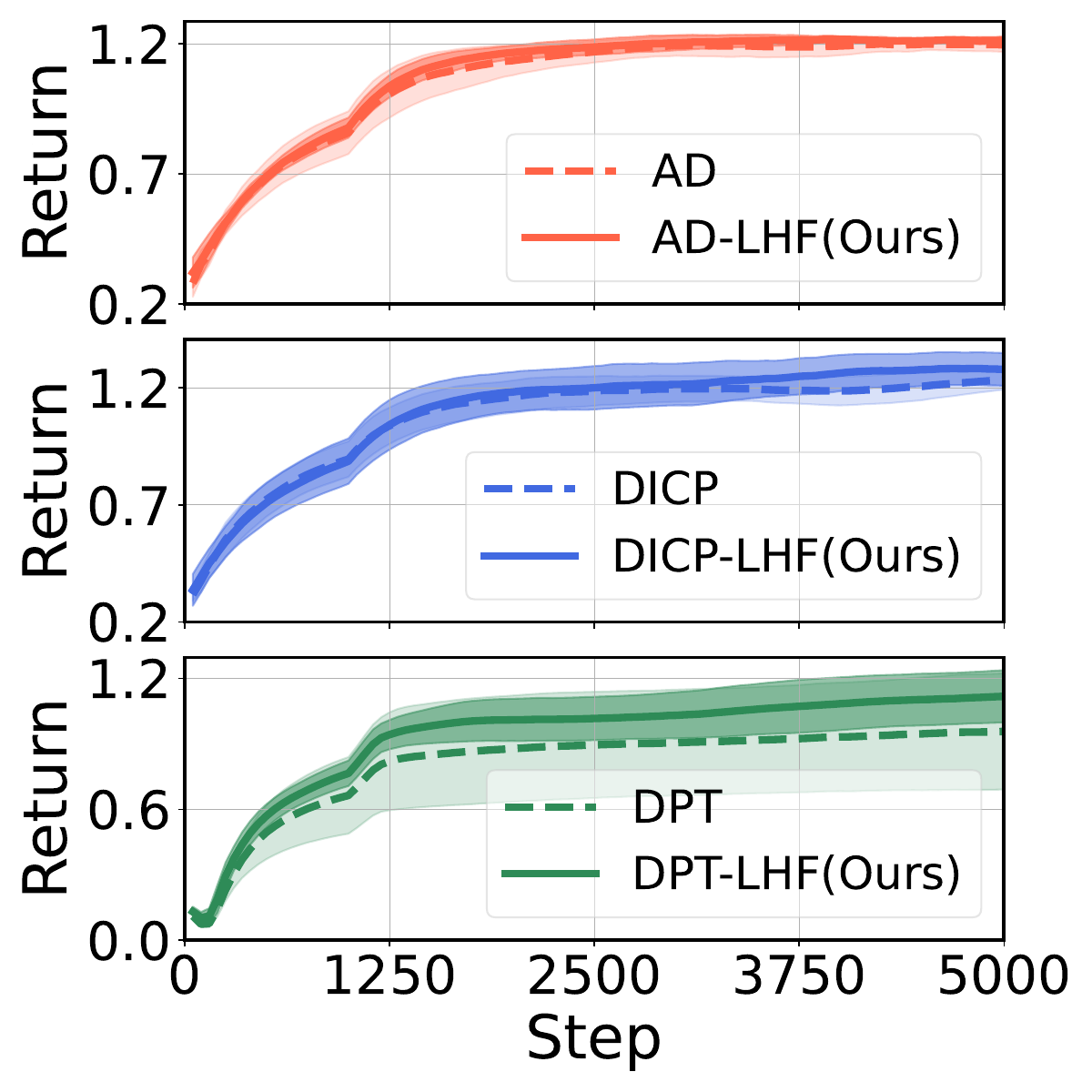}
	\end{minipage}
}
\caption{Learning curves of our LHF approach (solid lines) compared with original baselines (dashed lines) during the test. Each algorithm contains three independent runs with mean and standard
deviation. The backbone algorithms include AD (red), DICP (blue), and DPT (green).}
\label{fig_darkrooms_OriginaData_FullHistory}
\end{figure}

\begin{wraptable}{r}{0.5\textwidth}
    \centering
    \vspace{-5pt}
    \caption{Relative enhancement (\%) of LHF over the baselines, provided with the noisy dataset.}
    \label{tab_darkrooms_NoisyDataset}
    \begin{tabular}{cccc}    
        \toprule
        Task & AD & DICP & DPT \\
        \midrule
      \textit{DarkRoom} & \textbf{90.7} & 19.0 & 21.4 \\
      \textit{Darkroom-Permuted} & 5.5 & \textbf{9.5} & 4.0 \\
      \textit{Darkroom-Large} & 13.8 & \textbf{17.3} & 13.9 \\
      \textit{Dark Key-to-Door} & 1.2 & 4.2 & \textbf{10.1} \\
     \midrule
      Average & \textbf{27.8} & 12.5 & 12.4 \\
        \bottomrule
    \end{tabular}
    \vspace{-5pt}
\end{wraptable}

\textbf{Can LHF enhance ICRL?}
We collect the pretraining dataset as in Section~\ref{subsec_collect_filter_hist}, and implement the three backbone algorithms (AD, DICP, DPT) in four \textit{Darkroom}-type problems with and without LHF.
The numerical results are presented in Figure~\ref{fig_darkrooms_OriginaData_FullHistory} and Table~\ref{tab_darkrooms_OriginaData_FullHistory}.
All positive relative enhancement in Table~\ref{tab_darkrooms_OriginaData_FullHistory} implies the consistently improved performance of our LHF approach over the baselines across all backbone algorithms and problems. 
On average, AD, DICP, and DPT yield relative enhancement of $8.8\%$, $9.1\%$, and $11.9\%$, respectively. Notably, certain scenarios like using AD in \textit{Darkroom} and employing DPT in \textit{Darkroom-Large} can achieve more than $25\%$ performance enhancement.

\begin{figure}[ht]
\centering 
\setcounter{subfigure}{0}
\subfigure[Darkroom]
{
	\begin{minipage}{0.25\linewidth}
	\centering 
	\includegraphics[width=1.0\columnwidth]{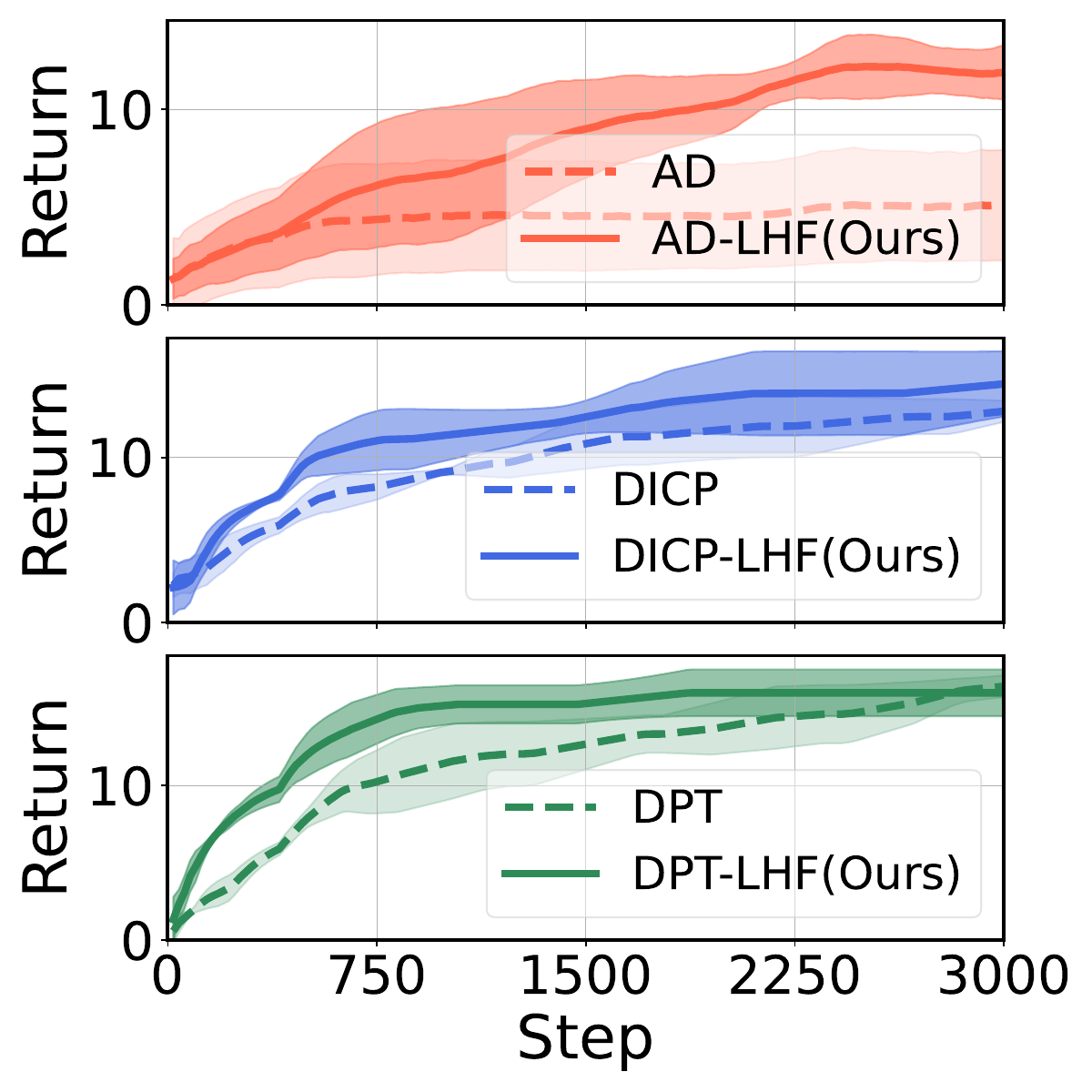}
	\end{minipage}
}
\!\!\!\!\!\!\!
\subfigure[Darkroom-Permuted]
{
	\begin{minipage}{0.25\linewidth}
	\centering 
	\includegraphics[width=1.0\columnwidth]{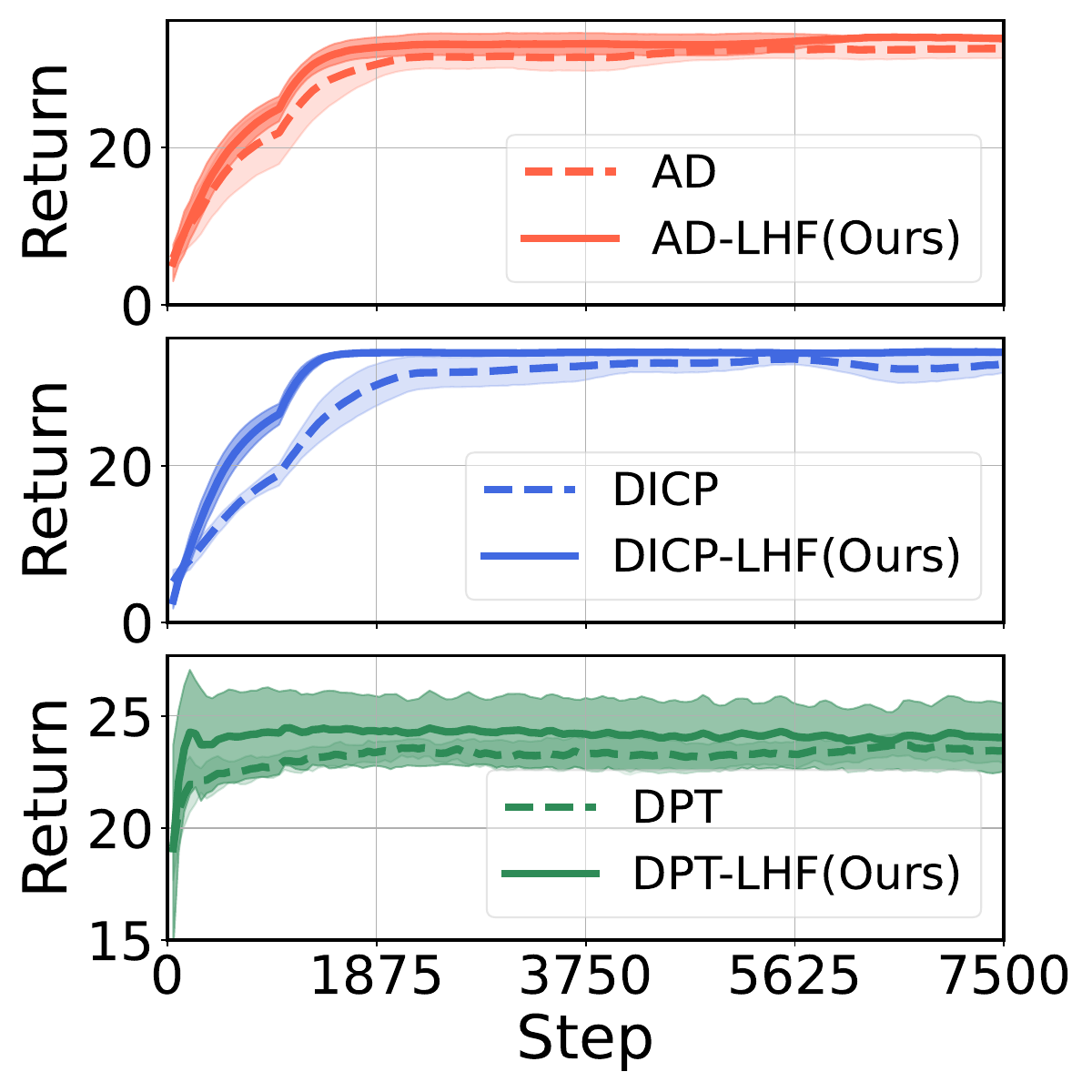}
	\end{minipage}
}
\!\!\!\!\!\!\!
\subfigure[Darkroom-Large]
{
	\begin{minipage}{0.25\linewidth}
	\centering 
	\includegraphics[width=1.0\columnwidth]{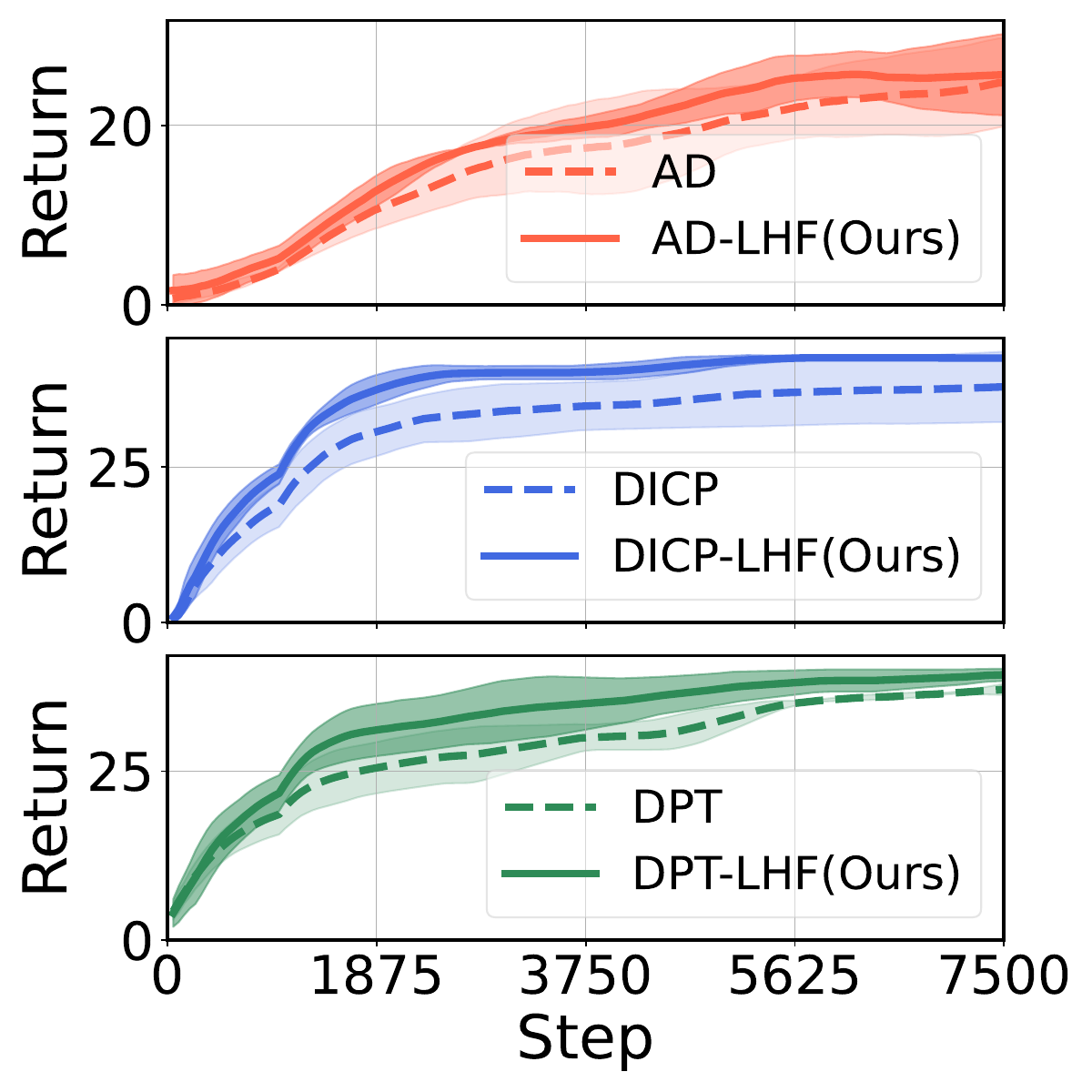}
	\end{minipage}
}
\!\!\!\!\!\!\!
\subfigure[Dark Key-to-Door]
{
	\begin{minipage}{0.25\linewidth}
	\centering 
	\includegraphics[width=1.0\columnwidth]{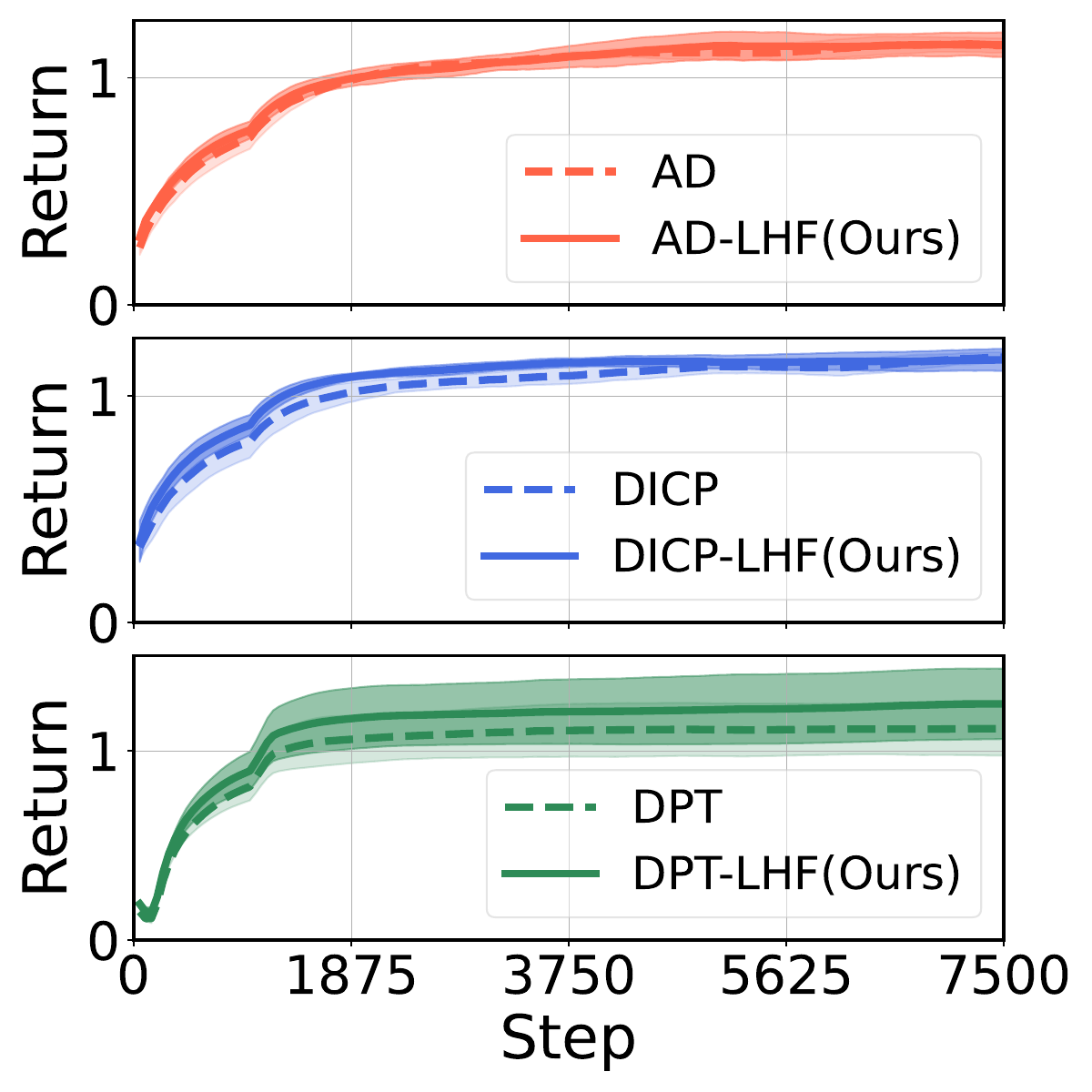}
	\end{minipage}
}
\caption{Learning curves of our LHF approach (solid lines) compared with original baselines (dashed lines) during the test. Each algorithm contains three independent runs with mean and std., provided with the noisy dataset. The backbone algorithms include AD (red), DICP (blue), and DPT (green).}
\label{fig_darkrooms_NoisyDataset}
\end{figure}

\textbf{Can LHF enhance ICRL given a noisy dataset?}
To validate the robustness of our LHF approach and to assess the significance of filtering learning histories, we now inject noises into the pretraining dataset. Concretely, the learning histories in the dataset are collected by $70\%$ PPO agents and $30\%$ random agents (executing uniform random actions).
The numerical results are presented in Figure~\ref{fig_darkrooms_NoisyDataset} and Table~\ref{tab_darkrooms_NoisyDataset}.
All positive relative enhancement in Table~\ref{tab_darkrooms_NoisyDataset} implies the consistently improved performance of LHF over the baselines across all backbone algorithms and problems, provided with the noisy dataset. 
On average, AD, DICP, and DPT yield relative enhancement of $27.8\%$, $12.5\%$, and $12.4\%$, respectively. It is worth highlighting that the noisy dataset (see Table~\ref{tab_darkrooms_NoisyDataset}) achieves an increased average relative enhancement than the dataset without the noises (see Table~\ref{tab_darkrooms_OriginaData_FullHistory}). In certain scenarios, such as employing AD in \textit{Darkroom}, performance enhancement can even exceed $90\%$. These results provide compelling evidence supporting the importance of filtering learning histories.

\textbf{Can LHF enhance ICRL given partial learning histories?}
Current ICRL algorithms require learning from sufficient improvements in order to distill the underlying improvement operator within the algorithm. Therefore, we now evaluate our LHF approach under a more challenging setting where only partial learning histories are provided. We select the first half ($50\%$) of learning histories from each environment in each problem, forming a new dataset with half learning histories only.
The numerical results are presented in Figure~\ref{fig_darkrooms_half_learn_hist} and Table~\ref{tab_darkrooms_half_learn_hist} (see Appendix~\ref{append_addition_result_half_learn_hist}).
All positive relative enhancement in Table~\ref{tab_darkrooms_half_learn_hist} except for the case of using DPT in \textit{Darkroom-Permuted} implies the consistently improved performance of our LHF approach over the baselines across most backbone algorithms and problems, provided with half learning histories.
On average, AD, DICP, and DPT yield relative enhancement of $11.2\%$, $4.9\%$, and $8.3\%$, respectively. Interestingly, the average performance enhancement of AD using half learning histories (see Table~\ref{tab_darkrooms_half_learn_hist}) is slightly higher than that using the complete learning histories (see Table~\ref{tab_darkrooms_OriginaData_FullHistory}). The certain scenario like employing AD in \textit{Darkroom-Large} demonstrates more than $22\%$ relative performance enhancement.


\begin{wraptable}{r}{0.31\textwidth}
    \centering
    \caption{Relative enhancement (\%) of LHF over baselines, provided with \textit{Meta-World-ML1}.}
    \label{tab_metaworld_ppo}
    \begin{tabular}{ccc}
    \toprule
    Task & AD & DICP \\
    \midrule
     \textit{Reach} & 2.4 & \textbf{64.6} \\
     \textit{Reach-Wall} & 5.9 & \textbf{141.3} \\
     \textit{Button-Press} & \textbf{13.4} & 2.4 \\
     \textit{Basketball} & 7.7 & \textbf{51.2} \\
     \textit{Door-Unlock} & 4.8 & \textbf{77.8} \\
     \textit{Push} & -0.4 & \textbf{18.9} \\
     \textit{Soccer} & \textbf{16.6} & 14.2 \\
     \textit{Hand-Insert} & \textbf{43.5} & -0.8 \\
     \midrule
     Average & 11.7 & \textbf{46.2} \\
    \bottomrule
    \end{tabular}
    \vspace{-5pt}
\end{wraptable}


\textbf{Can LHF enhance ICRL given lightweight models?}
We further investigate the performance of LHF provided with lightweight models. 
Given the hyperparameters of TMs as presented in Appendix~\ref{append_transformer_hyperparameter}, we select four representative hyperparameters: the number of attention layers ($4$), the number of attention heads ($4$), the embedding dimension ($32$), the intermediate size ($128$), and reduce each by half yielding $(2, 2, 16, 64)$.
The numerical results are presented in Figure~\ref{fig_darkrooms_lightweight_model} and Table~\ref{tab_darkrooms_lightweight_model} (see Appendix~\ref{append_addition_result_lightweight_model}).
All positive enhancement in Table~\ref{tab_darkrooms_lightweight_model}, except for the case of using DPT in \textit{Darkroom-Permuted}, implies the consistently improved performance of LHF over the baselines across most backbone algorithms and problems, provided with lightweight models.
On average, AD, DICP, and DPT yield relative enhancement of $12.6\%$, $13.0\%$, and $4.2\%$. The certain scenario, e.g., using DICP in \textit{Darkroom}, exhibits more than $28\%$ performance enhancement. Notably, average performance enhancements of AD and DICP using lightweight models (see Table~\ref{tab_darkrooms_lightweight_model}) exceed those achieved with heavyweight models (see Table~\ref{tab_darkrooms_OriginaData_FullHistory}), suggesting the possibility of overfitting in the latter.

\begin{figure}[ht]
\centering 
\setcounter{subfigure}{0}
\subfigure[Reach]
{
	\begin{minipage}{0.25\linewidth}
	\centering 
	\includegraphics[width=1.0\columnwidth]{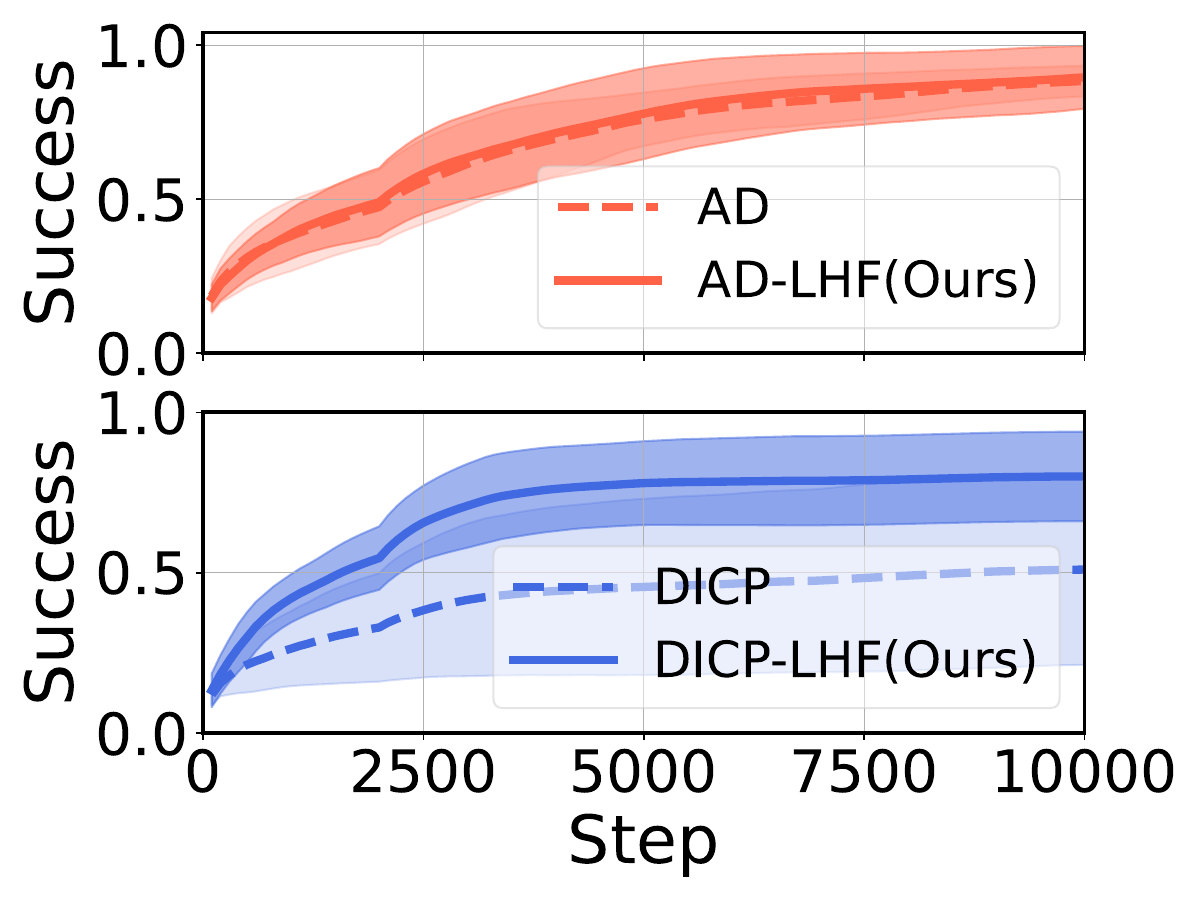}
	\end{minipage}
}
\!\!\!\!\!\!\!
\subfigure[Reach-Wall]
{
	\begin{minipage}{0.25\linewidth}
	\centering 
	\includegraphics[width=1.0\columnwidth]{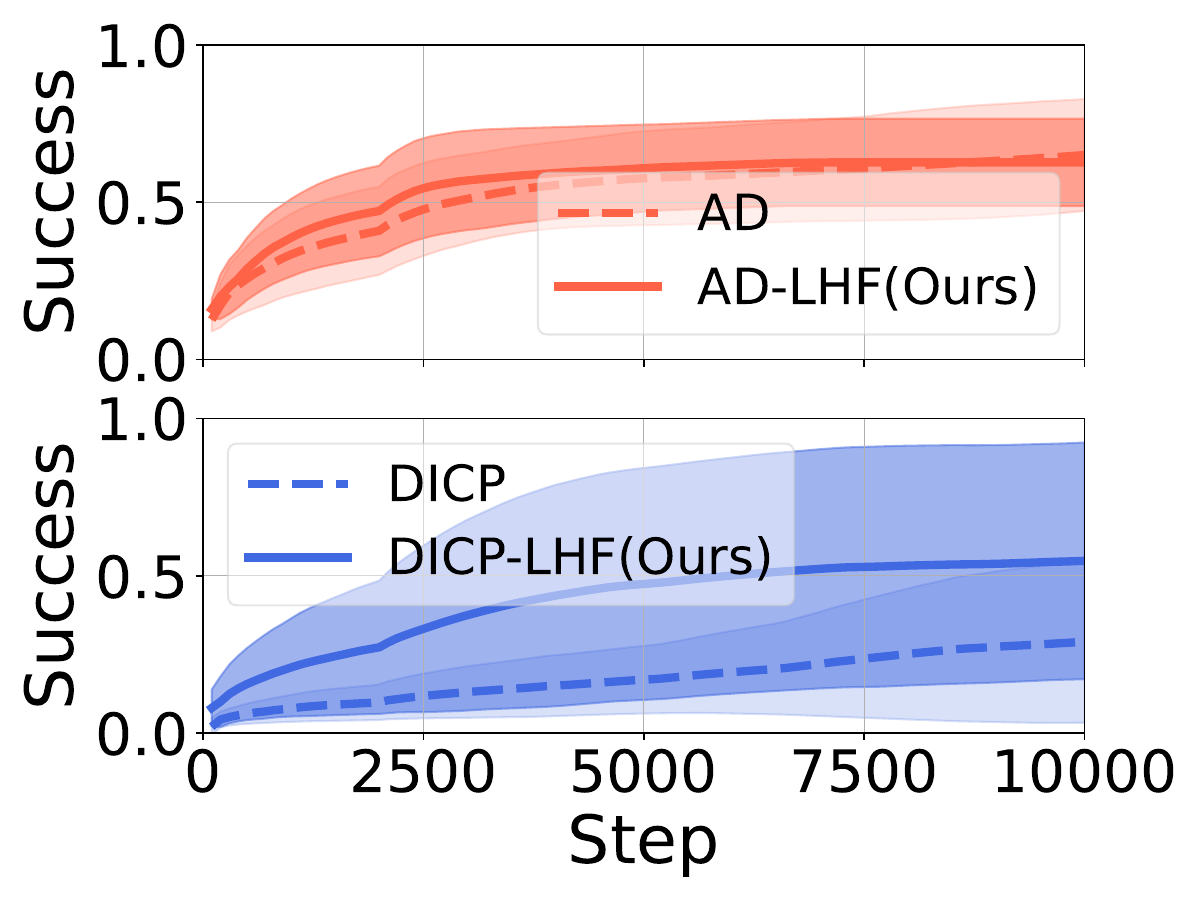}
	\end{minipage}
}
\!\!\!\!\!\!\!
\subfigure[Button-Press]
{
	\begin{minipage}{0.25\linewidth}
	\centering 
	\includegraphics[width=1.0\columnwidth]{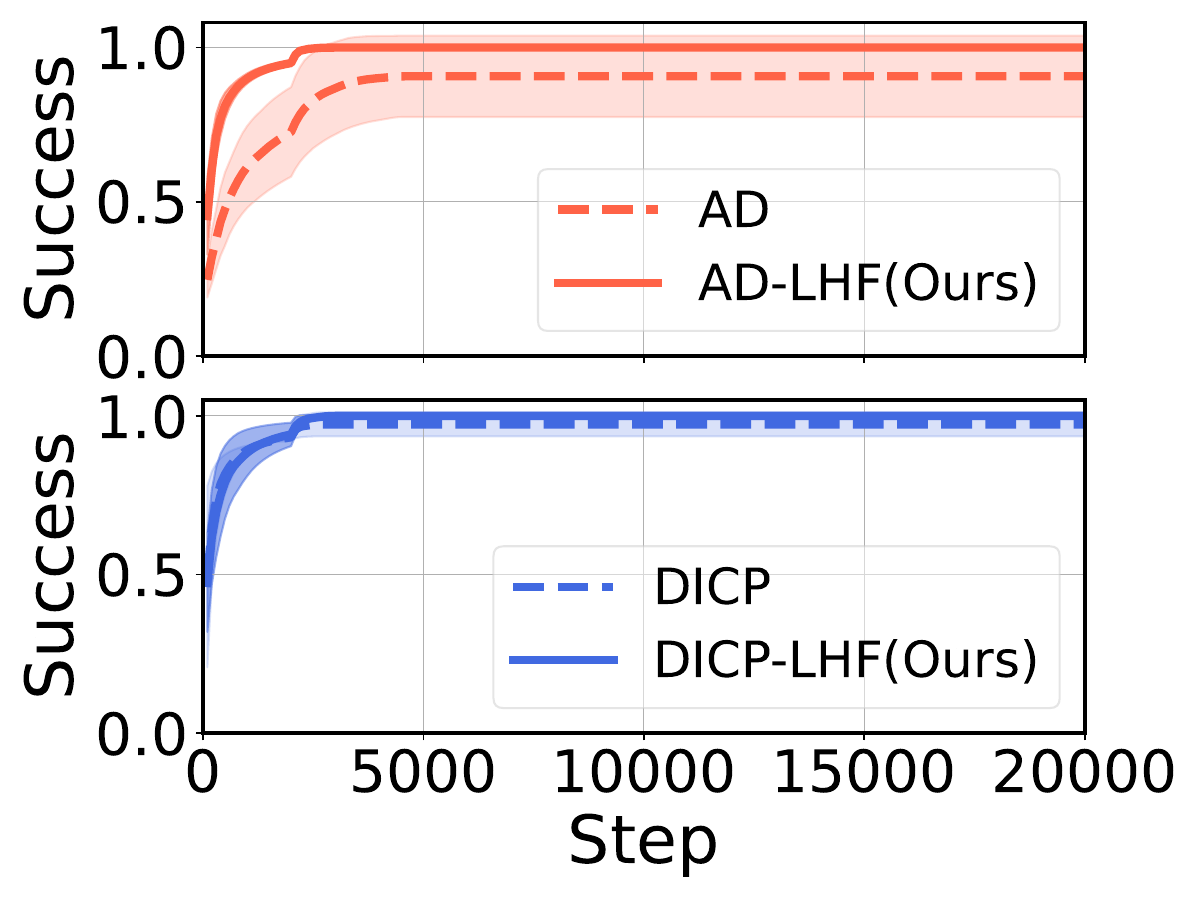}
	\end{minipage}
}
\!\!\!\!\!\!\!
\subfigure[Basketball]
{
	\begin{minipage}{0.25\linewidth}
	\centering 
	\includegraphics[width=1.0\columnwidth]{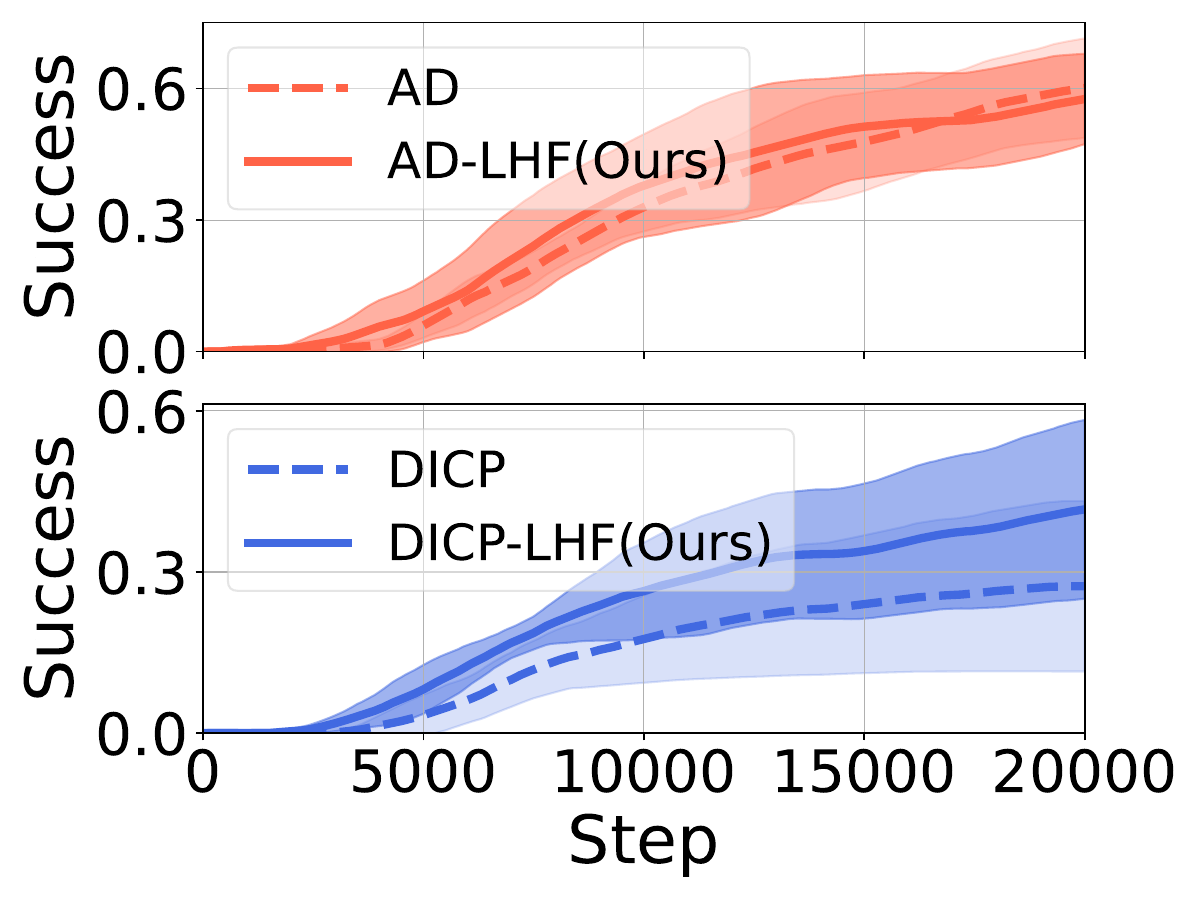}
	\end{minipage}
}
\\
\subfigure[Door-Unlock]
{
	\begin{minipage}{0.25\linewidth}
	\centering 
	\includegraphics[width=1.0\columnwidth]{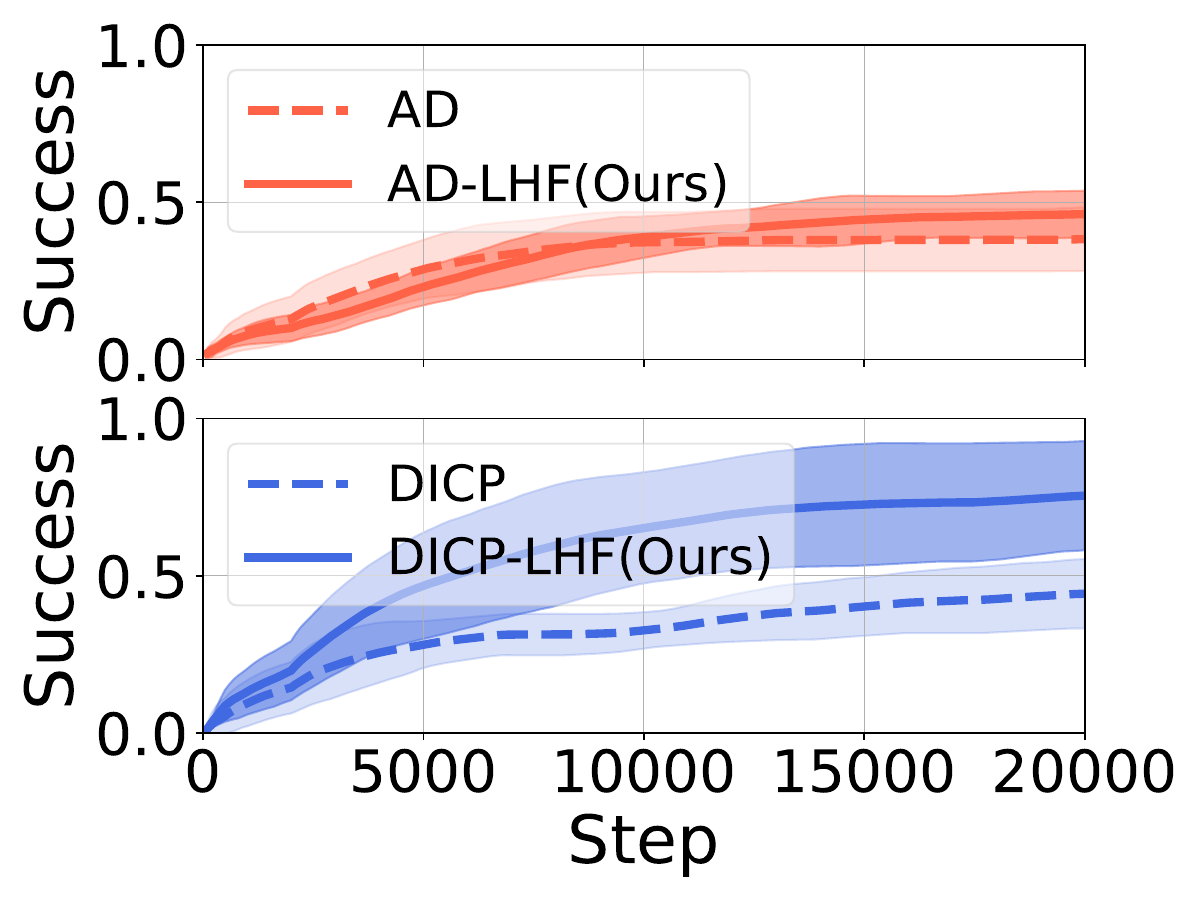}
	\end{minipage}
}
\!\!\!\!\!\!\!
\subfigure[Push]
{
	\begin{minipage}{0.25\linewidth}
	\centering 
	\includegraphics[width=1.0\columnwidth]{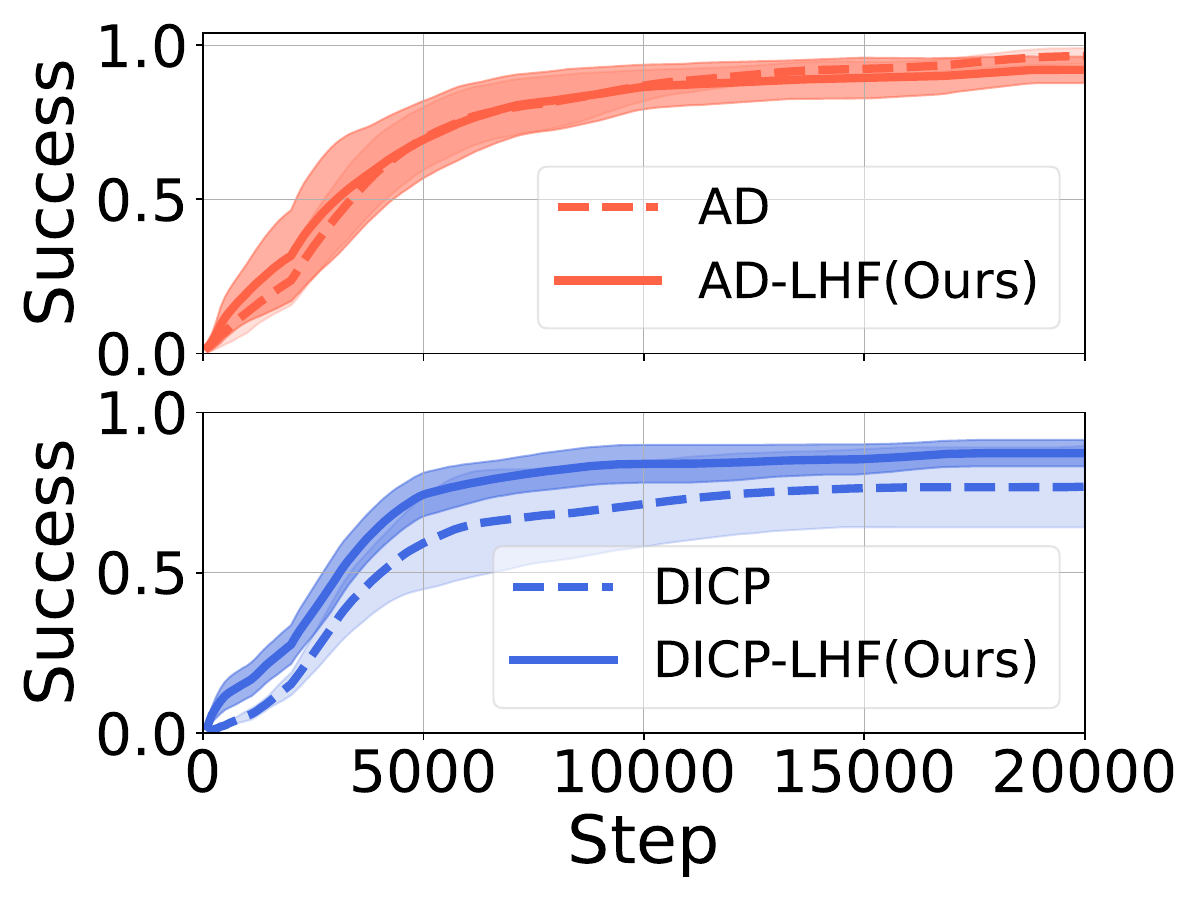}
	\end{minipage}
}
\!\!\!\!\!\!\!
\subfigure[Soccer]
{
	\begin{minipage}{0.25\linewidth}
	\centering 
	\includegraphics[width=1.0\columnwidth]{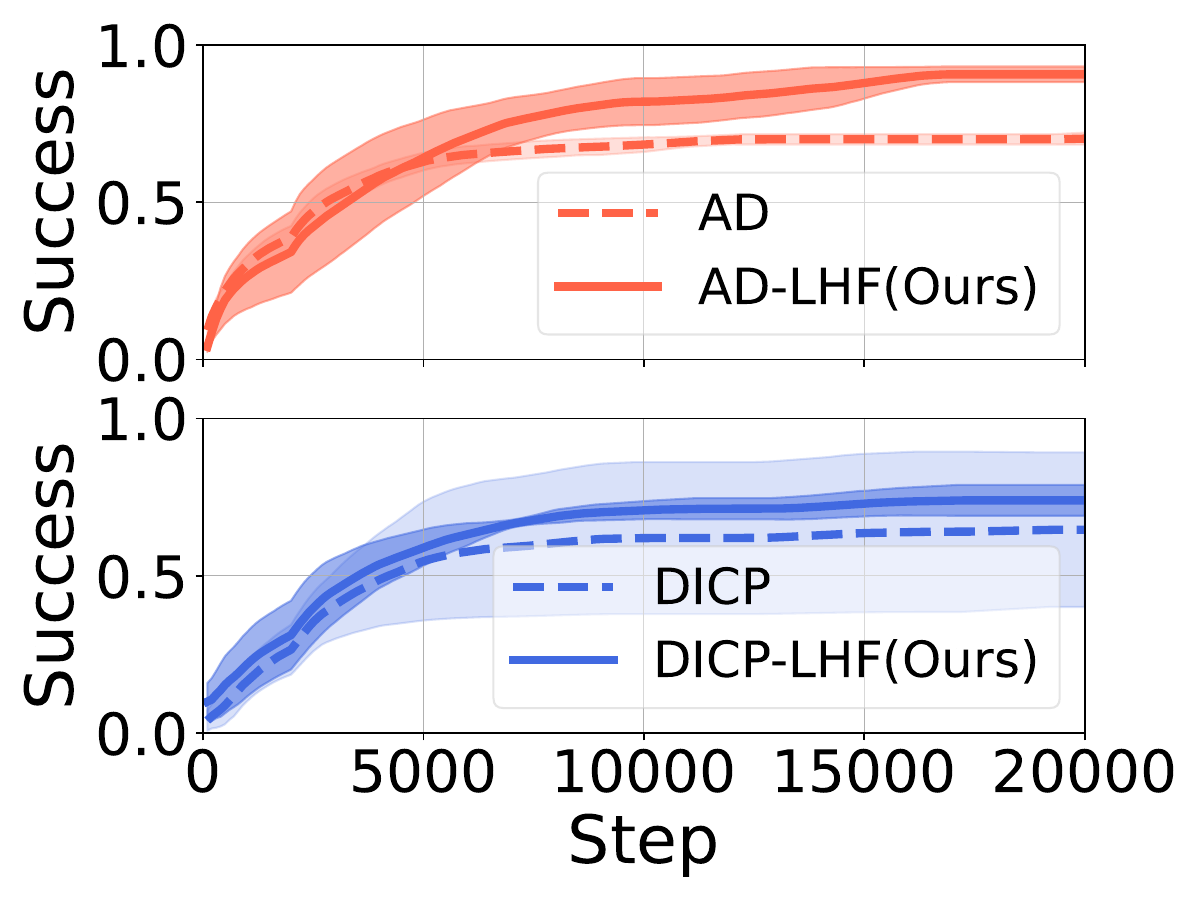}
	\end{minipage}
}
\!\!\!\!\!\!\!
\subfigure[Hand-Insert ]
{
	\begin{minipage}{0.25\linewidth}
	\centering 
	\includegraphics[width=1.0\columnwidth]{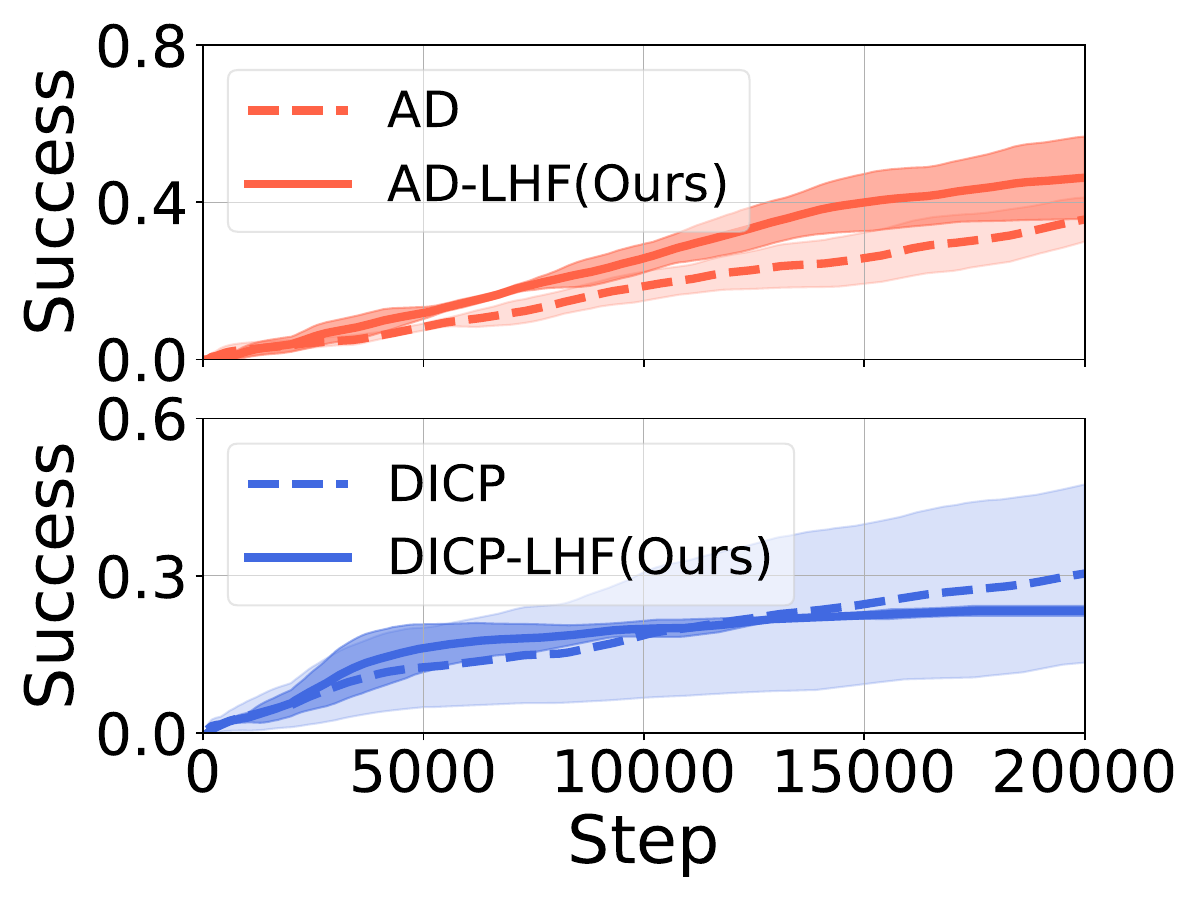}
	\end{minipage}
}
\caption{Learning curves of our LHF approach (solid lines) compared with original baselines (dashed lines) during the test. Each algorithm contains three independent runs with mean and std., provided with \textit{Meta-World-ML1} environments. The backbone algorithms include AD (red) and DICP (blue).}
\label{fig_metaworld_ppo}
\end{figure}

\textbf{Can LHF enhance ICRL for complex continuous robotic tasks?}
Having verified the superior performance of our LHF approach using the four discrete environments (\textit{Darkroom}-type), we further evaluate LHF using more complicated continuous tasks of robotic manipulations: \textit{Meta-World-ML1}.
As mentioned earlier, we consider only AD and DICP as backbones, since DPT requires optimal action labels that are not available in \textit{Meta-World-ML1}.
The numerical results are presented in Figure~\ref{fig_metaworld_ppo} and Table~\ref{tab_metaworld_ppo}.
All positive relative enhancement in Table~\ref{tab_metaworld_ppo} except the cases of using AD in \textit{Push} and employing DICP in \textit{Hand-Insert} implies the consistently improved performance of LHF over the baselines across most backbone algorithms and \textit{Meta-World-ML1} tasks. 
On average, AD and DICP yield relative enhancements of $11.7\%$ and $46.2\%$, respectively. Notably, the certain scenario such as using DICP in \textit{Reach-Wall} can achieve more than $141\%$ performance enhancement.

\subsection{Sensitivity Analysis}
\label{section_sensitivity}

\begin{figure}[ht]
\centering 
\setcounter{subfigure}{0}
\subfigure[AD with varying $\lambda$]
{
	\begin{minipage}{0.25\linewidth}
	\centering 
    \includegraphics[width=1.0\columnwidth]{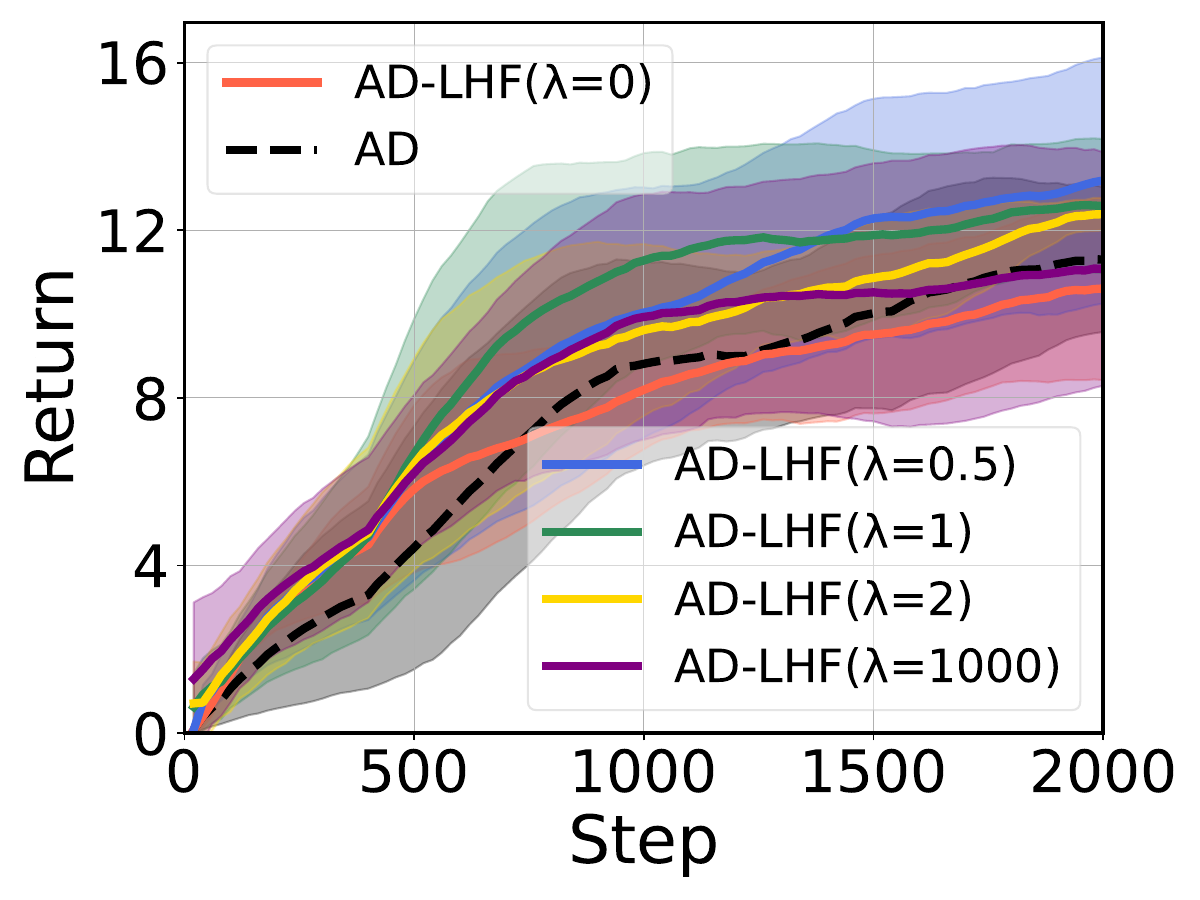}
	\end{minipage}
}
\!\!\!\!\!\!\!
\subfigure[DICP with varying $\lambda$]
{
	\begin{minipage}{0.25\linewidth}
	\centering 
	\includegraphics[width=1.0\columnwidth]{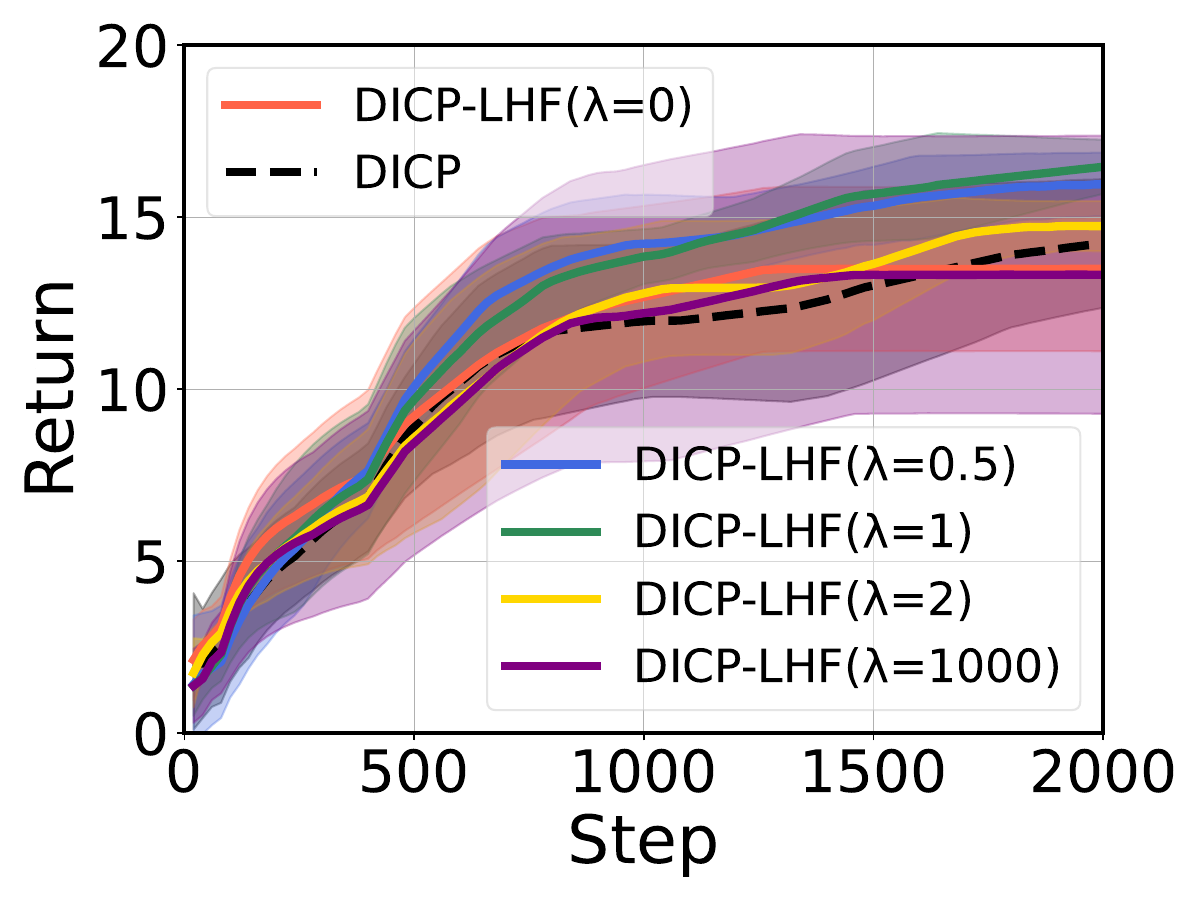}
	\end{minipage}
}
\!\!\!\!\!\!\!
\subfigure[AD with varying $\alpha$]
{
	\begin{minipage}{0.25\linewidth}
	\centering 
	\includegraphics[width=1.0\columnwidth]{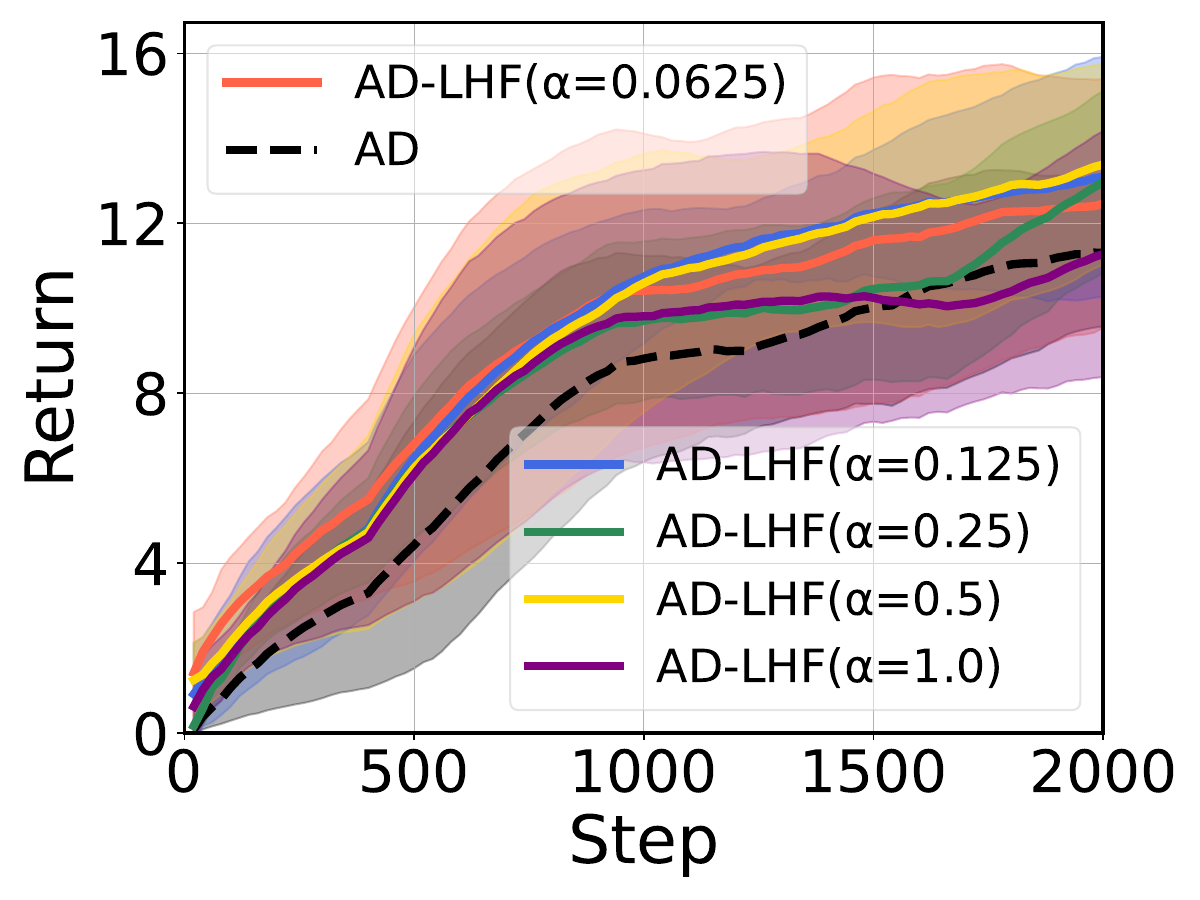}
	\end{minipage}
}
\!\!\!\!\!\!\!
\subfigure[DICP with varying $\alpha$]
{
	\begin{minipage}{0.25\linewidth}
	\centering 
	\includegraphics[width=1.0\columnwidth]{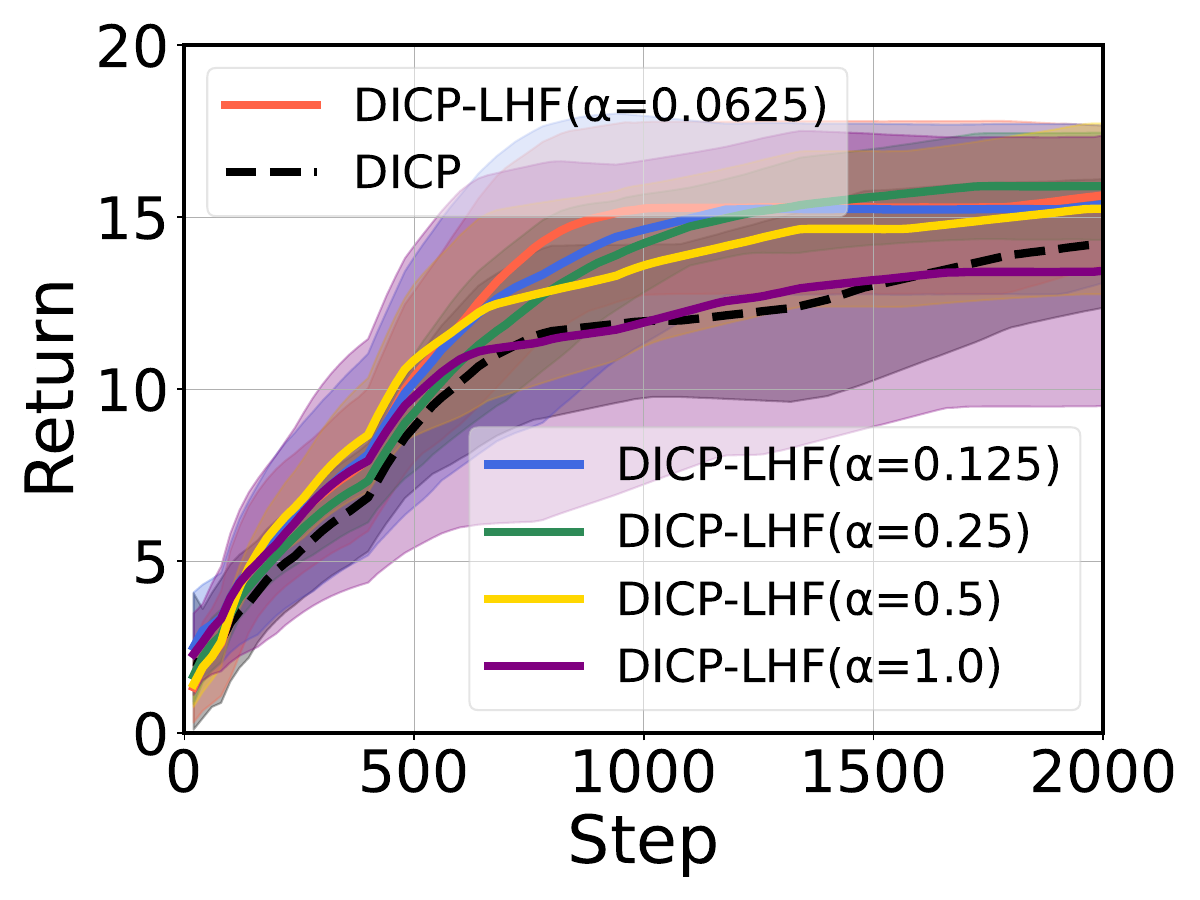}
	\end{minipage}
}
\caption{Learning curves of our LHF approach (solid lines) compared with original baselines (dashed lines) during the test. Each algorithm contains three independent runs with mean and std., provided with different stability coefficient $\lambda$ ((a) and (b)) and different temperature coefficient $\alpha$ in the Softmax sampling strategy ((c) and (d)). The backbone algorithms include AD and DICP.}
\label{fig_sensitivity}
\end{figure}

In all experiments thus far, we have used a fixed stability coefficient ($\lambda = 1$) and a fixed linear sampling strategy to evaluate the general performance and robustness of LHF.
However, as shown in \eqref{eqn_metric}, $\lambda$ governs the trade-off between the improvement and stability, both of which are critical for ICRL pretraining. Therefore, it is essential to investigate how varying $\lambda$ influences the performance of LHF.
We evaluate the backbone algorithms AD and DICP on the \textit{Darkroom} problem under varying stability coefficients $\lambda \in \{0, 0.5, 1, 2, 1000\}$. The corresponding numerical results are presented in Figure~\ref{fig_sensitivity}(a) and \ref{fig_sensitivity}(b).
Notably, for both algorithms, the settings $\lambda \in \{0.5, 1, 2\}$ consistently outperform the two extremes $\lambda \in \{0, 1000\}$, highlighting the significance of balancing the improvement and stability during the ICRL pertaining. Overall, even the worst-case performance under varying $\lambda$ remains comparable to the original baseline without LHF, demonstrating the robustness of our approach.

Now we turn our attention to the sampling strategies. In this sensitivity analysis, we adopt a Softmax sampling strategy with a temperature coefficient $\alpha$ as follows 
\begin{align}\label{eqn_softmax_weight_func}
    \mathcal{P}_{\bar{w}}(U(\ccalD_i^l)) = \frac{U_\text{soft}(\ccalD_i^l) - \min_{l \in [N_l]} U_\text{soft}(\ccalD_i^l)}{\max_{l \in [N_l]} U_\text{soft}(\ccalD_i^l) - \min_{l \in [N_l]} U_\text{soft}(\ccalD_i^l)}, \, \text{where} \,\,  U_\text{soft}(\ccalD_i^l) = \frac{e^{U(\ccalD_i^l) / \alpha}}{\sum_l e^{U(\ccalD_i^l) / \alpha}}
\end{align}

We also implement AD and DICP on \textit{Darkroom} under varying temperature coefficients $\alpha \in \{0.0625, 0.125, 0.25, 0.5, 1\}$. The numerical results are presented in Figure~\ref{fig_sensitivity}(c) and \ref{fig_sensitivity}(d).
Both algorithms exhibit the superiority of LHF compared to the baselines, with even the worst-case results remaining comparable. This validates the robustness of LHF with respect to the sampling strategies.

Notice that all preceding experiments consider PPO as the source RL algorithm. To verify the algorithm-agnostic nature of LHF, we now employ SAC~\cite{haarnoja2018soft} to collect learning histories across four \textit{Meta-World-ML1} tasks: \textit{Reach}, \textit{Button-Press}, \textit{Push}, \textit{Soccer}. The details of SAC algorithm is provided in Appendix~\ref{append_collect_history}.
The numerical results are presented in Figure~\ref{fig_sac_source_rl} and Table~\ref{tab_sac_source_rl} in Appendix~\ref{append_addition_result_sensitivity_sourceRL}. 
%
%
On average, AD and DICP yield relative enhancement of $44.0\%$ and $9.2\%$, respectively. LHF in the certain scenario such as employing AD in \textit{Reach} achieves more than $110\%$ performance enhancement. These findings empirically validate the robustness of LHF with respect to the source RL algorithm.

\section{Discussion}
\label{discussion_section}
In this work, we introduce the learning history filtering (LHF), a simple yet effective dataset preprocessing approach to enhance ICRL by addressing the issue of inheriting source suboptimality in the dataset. LHF operates by reweighting and filtering the learning histories according to their inherent improvement and stability, offering a general plug-in mechanism compatible with existing ICRL algorithms.
Through a series of evaluations on \textit{Darkroom}-type problems and \textit{Meta-World-ML1} robotic manipulation tasks, we demonstrate the superior performance and robustness of LHF aross various scenarios. The performance enhancement is even more obvious on the noisy datasets, further underscoring the significance of filtering suboptimal histories.
Our findings also demonstrate the importance and success of data-centric interventions in advancing the performance of ICRL.

\textbf{Limitations and Future Work.}
Our LHF approach is inspired by the WERM schema, which is natural and intuitive from an optimization perspective. However, it remains an open and interesting direction in the future to theoretically characterize the filtering mechanism in the specific context of ICRL with respect to e.g., generalization error~\cite{lin2023transformers}.
Moreover, existing ICRL methods fall into the category of unconstrained RL, which remains inadequate for safety-critical applications. Future work could incorporate an extra cost function, analogous to reward, to enable the in-context safe RL.


\clearpage

\section*{Acknowledgement}
The authors would like to thank Jaehyeon Son for the valuable discussions as well as the open-source implementations of DICP.

\bibliography{neurips_2025}
\bibliographystyle{unsrt}

\clearpage

\appendix

\section{Implementation and Experiment Details}




\subsection{Collecting Learning Histories}
\label{append_collect_history}

In this work, we employ the Stable Baselines 3 (SB3) implementations of Proximal Policy Optimization (PPO) \citep{schulman2017proximal} and Soft Actor–Critic (SAC) \citep{haarnoja2018soft} to generate learning histories for ICRL pre-training. PPO is an on-policy algorithm that stabilises updates with a clipped surrogate objective, whereas SAC is an off-policy, entropy-regularised actor–critic method that encourages exploration via a maximum-entropy objective. SB3 provides well-tested PyTorch versions of both algorithms under a uniform API, which facilitates reproducibility. 
Following DICP \cite{son2025distilling}, we summarize the key hyperparameters in Table \ref{tab:hyp-ppo-sac} while the remaining hyperparameters are kept at default.

\newcommand{\adjvcenter}[1]{%
  \raisebox{-.55\height}[0pt][0pt]{#1}%
}

\begin{table}[h]
\caption{Key hyperparameters for PPO and SAC.}
\label{tab:hyp-ppo-sac}
\scriptsize
\centering
\setlength{\tabcolsep}{3pt}
\begin{tabular}{l c c c c c c}
\toprule
\multirow{2}{*}{\adjvcenter{\textbf{Hyperparameter}}} &
\multicolumn{5}{c}{\textbf{PPO}} & \textbf{SAC}\\
\cmidrule(lr){2-6}\cmidrule(lr){7-7}
& \textit{Darkroom} & \textit{Darkroom-Permuted} &
  \textit{Darkroom-Large} & \textit{Dark Key-to-Door} &
  \textit{Meta-World-ML1} & \textit{Meta-World-ML1}\\
\midrule
batch size              & 50  & 50  & 50  & 100 & 200 & 128\\
discount factor         & 0.99& 0.99& 0.99& 0.99& 0.99& 0.99\\
source learning rate    & $3\times10^{-4}$ & $3\times10^{-4}$ &
                          $3\times10^{-4}$ & $3\times10^{-4}$ &
                          $3\times10^{-4}$ & $3\times10^{-4}$\\
\# of processes         & 8   & 8   & 8   & 8   & 8   & 8\\
\# of learning histories& 100 & 100 & 100 & 100 & 100 & 100\\
total transitions       & $1\times10^{5}$ & $1\times10^{5}$ & $1\times10^{5}$ &
                          $1\times10^{5}$ & $1\times10^{6}$ & $1\times10^{6}$\\
\bottomrule
\end{tabular}
\end{table}

\subsection{Backbone ICRL Algorithms}
\label{append_backboneICRLalgo}

\paragraph{Algorithm Distillation (AD).}
AD \citep{laskin2022context} is an in-context RL framework for transforming the training process of a source RL algorithm into a single in-context policy. Concretely, AD first collects learning histories from an RL algorithm deployed on a large substantial amount of tasks. Each learning history is a multi-episode record of transitions $\bigl(s_t^{(i)},\,a_t^{(i)},\,r_t^{(i)}\bigr)$, capturing how the source algorithm explores and improves its policy. A TM is then trained, via supervised learning, to predict the source algorithm’s action $a_t^{(i)}$ from the preceding history
\begin{equation}
\label{eq:ad}
\gL_{\mathrm{AD}}(\theta)
=
- \sum_{i \in [N]}\!
  \sum_{t \in [T]}
  \log
  M_{\theta}\!\Bigl(
      a_t^{(i)}
      \,\Bigm|\,
      \ccalC_{t-1}^{(i)},\,  
      s_t^{(i)}
  \Bigr),
\end{equation}
 where $\ccalC_{t-1}^{(i)}$ denotes the context of the $i$-th learning history up to step $t{-}1$, and $M_\theta(\cdot\mid\cdot)$ is the model’s predicted action distribution. Once trained, the TM can be deployed \emph{without} any parameter updates on new tasks, adapting online by conditioning on its own growing history. By imitating \emph{entire learning sequences} rather than a single policy snapshot, AD yields an in-context learner that inherits effective exploration and credit-assignment strategies from its source algorithm.

\paragraph{Decision Pretrained Transformer (DPT).}
DPT~\cite{lee2024supervised} is another in-context RL method that pre-trains a TM to predict optimal (or near-optimal) actions given a sampled query state and context during the ICRL pretraining. Throughout this paper, we consider the query state and context in DPT to be sampled from the learning histories, as introduced in the original DPT paper~\cite{lee2024supervised}.
In practice, DPT requires either an oracle or a well-trained expert policy that can generate high-quality actions for labeling all pretraining tasks.
Formally, the DPT objective can be written as
\begin{equation}
    \label{eq:dpt2}
    \gL_{\mathrm{DPT}}(\theta)
    \;=\;
- \sum_{i \in [N]}\!
  \sum_{t \in [T]}
      \log\,
      M_{\theta}\!\Bigl(
        a_t^{(i), \star}
        \;\Bigm|\;
        \ccalC_{t-1}^{(i)},\,  
         s_t^{(i)}
      \Bigr),
\end{equation}
where $a_t^{(i), \star}$ denotes the optimal action for the state $s_t^{(i)}$ in the underlying MDP. 
Under mild conditions on the task distribution and sufficient model capacity, DPT can approximate a Bayesian posterior over tasks, thus emulating posterior-sampling-style updates in context. Consequently, it is able to learn efficient strategies for online exploration and offline decision-making purely through a supervised objective.

\paragraph{Distillation for In-Context Planning (DICP).}  
DICP~\citep{son2025distilling} is a \emph{model-based} extension of ICRL, built on AD~\citep{laskin2022context} and DPT~\citep{lee2024supervised}. 
Instead of only predicting an action for each in-context step, DICP also learns a \emph{dynamics model} in-context, enabling the agent to simulate future transitions before acting.
Formally, a TM is pretrained to model not only $a_t$ (action), but also $(r_t,\,s_{t+1}, R_t)$ (reward, next state, return-to-go), yielding an objective as follows
\begin{equation}
\label{eq:dicp}
\begin{aligned}
\gL_{\mathrm{DICP}}(\theta)
= &~
\underbrace{
- \sum_{i \in [N]}\!
  \sum_{t \in [T]}
  \log 
  M_{\theta}\Bigl(
    a_{t}^{(i)}
    \,\Bigm|\,
    \ccalC_{t-1}^{(i)},\
    s_{t}^{(i)}
  \Bigr)
}_{\text{\normalsize imitation of the source algorithm}}
\\[0.3em]
& \quad
+ \;
\xi \;
\underbrace{
  \Big(- \sum_{i \in [N]}\!
  \sum_{t \in [T]}
  \log 
  W_{\theta}\Bigl(
    r_{t}^{(i)},\,s_{t+1}^{(i)},\, R_t^{(i)}
    \,\Bigm|\,\ccalC_{t-1}^{(i)},\,s_{t}^{(i)}, \,
    a_{t}^{(i)}
  \Bigr)\Big)
}_{\text{\normalsize modeling dynamics}}
\,,
\end{aligned}
\end{equation}
where \(\xi\) is a hyperparameter that balances algorithm imitation and dynamics modeling.
Once pretrained, at test time DICP applies \emph{planning} (e.g.\ beam or greedy search) over multiple simulated trajectories drawn from $W_{\theta}(\cdot)$ to choose an action that maximizes predicted rewards. By leveraging the learned dynamics model in-context, DICP enables the improvement of ICRL performance especially when the source algorithm exhibits suboptimal behaviors. Thus, compared to prior ICRL methods, DICP enables more deliberate decision-making through model-based search, further enhancing the sample efficiency and adaptability.

\subsection{Transformer Models}
\label{append_transformer_hyperparameter}

TMs employed in this work are based on the open-source \emph{TinyLlama} framework~\citep{zhang2024tinyllama}, 
a lightweight yet powerful model designed for efficient large language model variants. 
Our experiments cover four discrete tasks (\textit{Darkroom}, \textit{Darkroom-Permuted}, \textit{Darkroom-Large}, \textit{Dark Key-to-Door}), which we collectively refer to as ``Gridworld'', and one continuous robotic manipulation benchmark (\textit{Meta-World-ML1}), denoted as ``Metaworld''. 
Table~\ref{tab:hyp-tf} provides the specific hyperparameter configurations we consider for AD, DICP, and DPT in these respective settings.
\begin{table}[h]
\caption{Key hyperparameters for discrete tasks (\textit{Gridworld}) and continuous robotic manipulation tasks (\textit{Metaworld}).}
\label{tab:hyp-tf}
\centering
\begin{tabular}{lcc}
\toprule
\textbf{Hyperparameter} & \textbf{Gridworld (AD, DICP, DPT)} & \textbf{Metaworld (AD, DICP)} \\
\midrule
attention dropout \& dropout   & 0.1           & 0.1 \\
$\beta_1$             & 0.9           & 0.9 \\
$\beta_2$             & 0.99          & 0.99\\
intermediate size    & 128           & 128 \\
learning rate         & $1\times10^{-2}$ & $1\times10^{-2}$ \\
embedding dimension          & 32            & 32  \\
\# of heads           & 4             & 4   \\
\# of layers          & 4             & 4   \\
optimizer             & AdamW         & AdamW \\
scheduler             & cosine decay  & cosine decay \\
weight decay         & 0.01          & 0.01 \\
\bottomrule
\end{tabular}
\end{table}

\subsection{Complete Process of ICRL via Learning History Filtering (LHF)}
\label{append_complete_pseudocode}
Below we provide a pseudo-code description of the full procedure for
applying learning history filtering (LHF) to ICRL. In
Algorithm~\ref{alg_whole}, we first filter the collected learning histories (lines~2--17), producing a refined pretraining dataset $\ccalD_{\text{LHF}}$. We then follow the standard pretraining and test processes (lines~18--28).
\begin{algorithm}
\caption{In-Context Reinforcement Learning via Learning History Filtering}         
\label{alg_whole}
\begin{algorithmic}[1]

        \STATE \textbf{Require:} Pretraining dataset $\{\ccalD_i^l\}$ with $i \in [N_i], l \in [N_l]$, empty LHF dataset $\ccalD_\text{LHF}$, initial model parameters $\theta$, test environment distribution $\ccalT_{\text{test}}$, number of test episodes $N_E$

        \STATE \green{\textbf{// Dataset Preprocessing}}
        \FOR{$i$ in $[N_i]$}

            \STATE Let $\ccalD_i' = \emptyset$
            \WHILE{$|\ccalD_i'| < |\ccalD_i|$}
                    
                \FOR{$l$ in $[N_l]$}

                    \STATE Compute the unified metric $U(\ccalD_i^l)$ by \eqref{eqn_metric}
                    \STATE Compute the weighted probability $\mathcal{P}_{\bar{w}}(U(\ccalD_i^l))$ for the learning history $\ccalD_i^l$ by \eqref{eqn_linear_weight_func}
            
                    \STATE Sample a uniform random variable $v \sim \ccalU[0, 1]$

                    \STATE Add the learning history $\ccalD_i^l$ to $\ccalD_i'$ \textbf{if} $v \leq \mathcal{P}_{\bar{w}}(U(\ccalD_i^l))$
                    
                    \IF{$|\ccalD_i'| = |\ccalD_i|$}
                    \STATE break
                    \ENDIF
                \ENDFOR
            \ENDWHILE
 
            \STATE $\ccalD_\text{LHF} \leftarrow \ccalD_\text{LHF} \cup \ccalD_i'$
        \ENDFOR

        \STATE \green{\textbf{// Pretraining}}
        \WHILE{not converged}
            \STATE Sample $(\ccalC, s_q, a_l)$ from the LHF dataset $\ccalD_\text{LHF}$ and predict actions by $M_\theta(\cdot  | \ccalC, s_q)$
            \STATE Compute the loss in \eqref{eqn_icrl_obj_weighted_data} with respect to the action label $a_l$ and backpropagate to update $\theta$.
        \ENDWHILE

        \STATE \green{\textbf{// Test}}
        \STATE Sample unseen test environments $\tau \sim \ccalT_{\text{test}}$ and initialize empty context $\ccalC = \{\}$

        \FOR{$n$ in $[N_E]$}
                \STATE Deploy $M_\theta$ by sampling $a_t \sim M_{\theta}(\cdot  \mid \ccalC, s_t)$ at time step $t$
                \STATE Add $(s_0, a_0, r_0, \ldots)$ to $\ccalC$
        \ENDFOR
\end{algorithmic}
\end{algorithm}

\subsection{Environmental Setup}
\label{append_env_setup}

\paragraph{\textit{Darkroom}.}
\textit{Darkroom} is a two-dimensional navigation task with discrete state and action spaces. The room consists of $9 \times 9$ grids, with the agent reset in the middle of the room and an unknown goal randomly placed at any of these grids. The agent can select $5$ actions: go up, go down, go left, go right, or stay. The horizon length of \textit{Darkroom} is $20$. One challenge of this task arises from its sparse reward structure, i.e., the agent receives a reward of $1$ solely upon reaching the goal, and $0$ otherwise. 
Given $9 \times 9 = 81$ available goals, we randomly select $73$ of these goals ($\sim 90\%$) for pretraining and reserve the remaining $8$ goals ($\sim 10\%$ and unseen during pretraining) for test.

\paragraph{\textit{Darkroom-Permuted}.}
\textit{Darkroom-Permuted} is a variant of \textit{Darkroom} with the same state space and reward structure, with the agent reset in a fixed corner of the room and the goal placed in the opposite corner. In this problem, the action space undergoes a random permutation, yielding $5! = 120$ distinct tasks with each defined by a unique permutation of the action space. The horizon length of \textit{Darkroom-Permuted} is $50$. We randomly select $108$ tasks ($90\%$) for pretraining and reserve the remaining $12$ tasks ($10\%$ and unseen during pretraining) for test.

\paragraph{\textit{Darkroom-Large}.}
\textit{Darkroom-Large} adopts the same setup as in \textit{Darkroom}, yet with an expanded state space of $15 \times 15$ and a longer horizon of $50$. Thus, the agent must explore the room more thoroughly due to the sparse reward setting, rendering this task more challenging than \textit{Darkroom}. We still consider $90\%$ of $15 \times 15 = 225$ available goals for pretraining and the remaining unseen $10\%$ goals for test.

\paragraph{\textit{Dark Key-to-Door}.}
\textit{Dark Key-to-Door} also adopts the same setup as in \textit{Darkroom}, yet with an extra ``key'' positioned in any of  the grids. The agent must locate the key before reaching the door (goal). In this setting, it receives a one-time reward of $1$ upon finding the key, followed by an additional one-time reward of $1$ upon reaching the door, yielding a maximum return of $2$ within this environment.
Given $81 \times 81 = 6561$ available tasks by distinct positions of the key and door, we randomly select $6233$ tasks ($\sim 95\%$) for pretraining and reserve the remaining $328$ tasks ($\sim 5\%$) for test.

\paragraph{\textit{Meta-World-ML1}.}
\textit{Meta-World ML1}~\cite{yu2020meta} focuses on a single robotic manipulation task at a time, with $50$ predefined seeds each for the pretraining and test. These seeds correspond to different initializations of the object, goal, and agent. The agent is trained with varying goal configurations, and tested on new (unseen) goals. 
In this work, we focus on $8$ distinct tasks: \textit{Reach, Reach-Wall, Button-Press, Basketball, Door-Unlock, Push, Soccer, Hand-Insert}, each with a horizon of $100$ steps.

\section{Additional Experimental Results}

\subsection{ICRL with partial learning histories}
\label{append_addition_result_half_learn_hist}

\begin{figure}[ht]
\centering 
\setcounter{subfigure}{0}
\subfigure[Darkroom]
{
	\begin{minipage}{0.25\linewidth}
	\centering 
	\includegraphics[width=1.0\columnwidth]{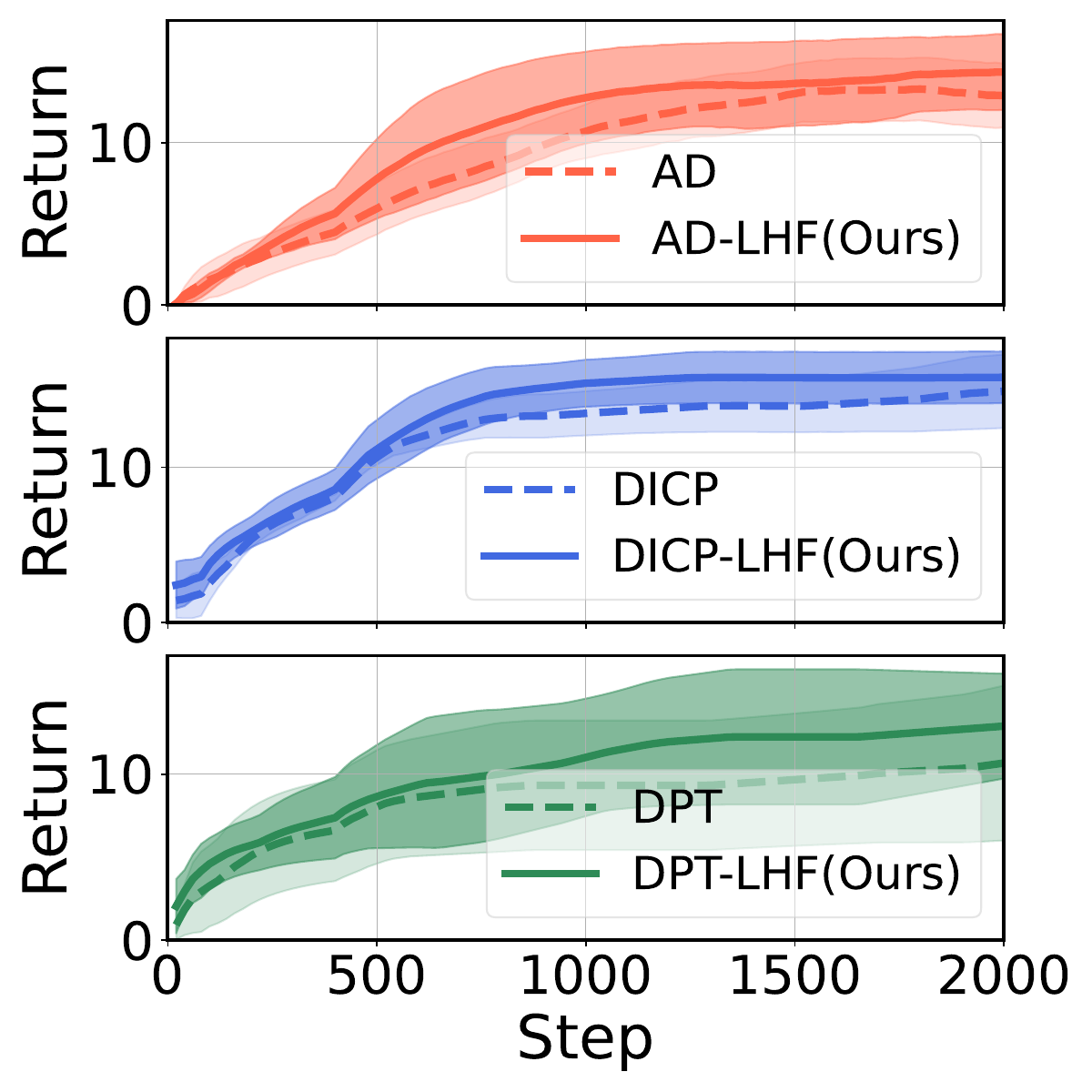}
	\end{minipage}
}
\!\!\!\!\!\!\!
\subfigure[Darkroom-Permuted]
{
	\begin{minipage}{0.25\linewidth}
	\centering 
	\includegraphics[width=1.0\columnwidth]{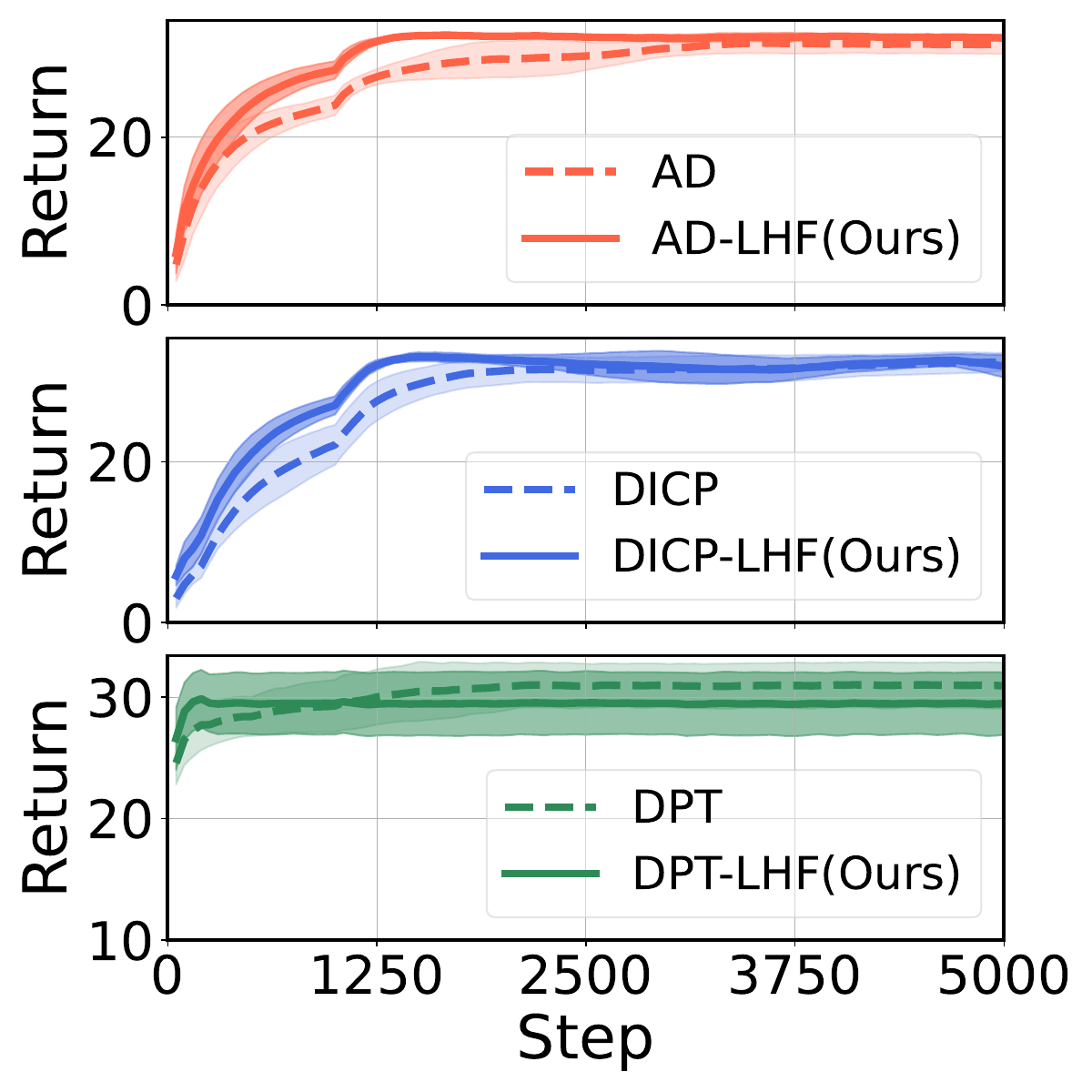}
	\end{minipage}
}
\!\!\!\!\!\!\!
\subfigure[Darkroom-Large]
{
	\begin{minipage}{0.25\linewidth}
	\centering 
	\includegraphics[width=1.0\columnwidth]{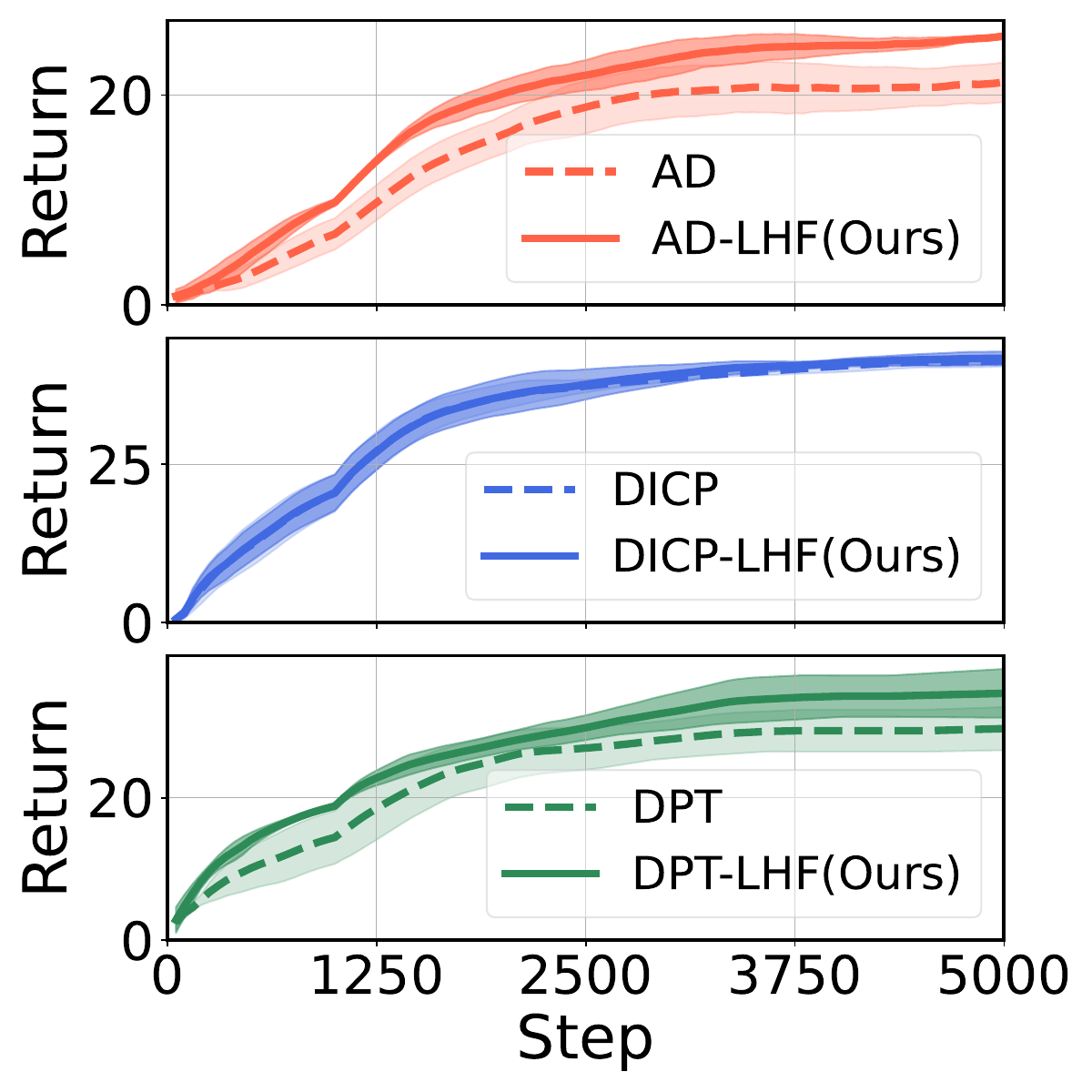}
	\end{minipage}
}
\!\!\!\!\!\!\!
\subfigure[Dark Key-to-Door]
{
	\begin{minipage}{0.25\linewidth}
	\centering 
	\includegraphics[width=1.0\columnwidth]{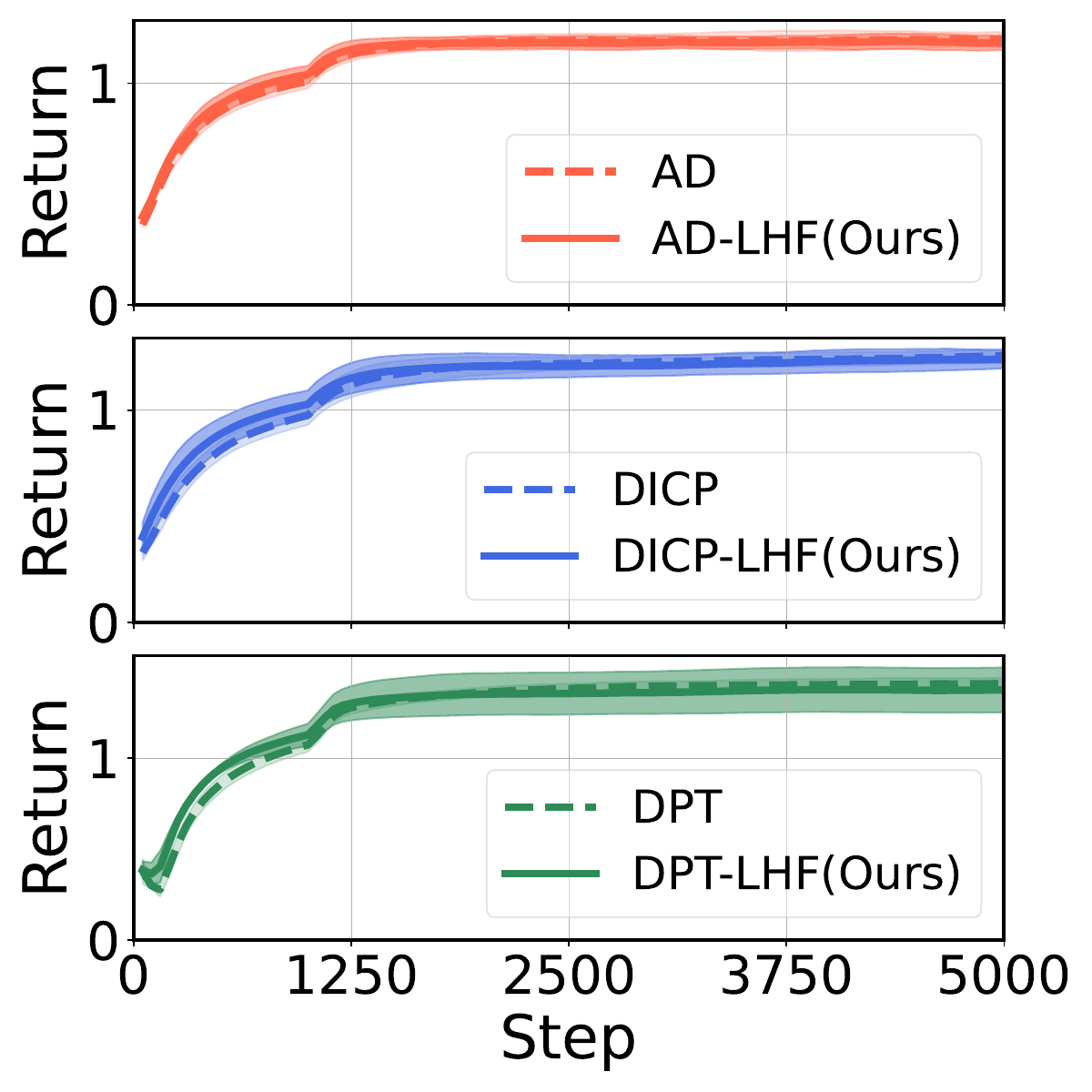}
	\end{minipage}
}
\caption{Learning curves of our LHF approach (solid lines) compared with original baselines (dashed lines) during the test. Each algorithm contains three independent runs with mean and standard deviation, provided with half learning histories. The backbone algorithms include AD (red), DICP (blue), and DPT (green).}
\label{fig_darkrooms_half_learn_hist}
\end{figure}

Section~\ref{exp_numerical_result} presents the challenging scenario in which only the first \(50\%\) of each PPO learning history is retained.  
Discarding the last half of each learning history diminishes the improvement signal and shortens the credit-assignment horizon, yet LHF still surpasses the unfiltered baselines in nearly every algorithm–task combination. 
Complete results are reported in
Table~\ref{tab_darkrooms_half_learn_hist}, and the associated learning curves are depicted in Figure~\ref{fig_darkrooms_half_learn_hist}.

\begin{table}[ht]
    \caption{Relative enhancement (\%) of our LHF approach over the baselines, provided with half learning histories. Backbone algorithms: AD, DICP, DPT.}
    \label{tab_darkrooms_half_learn_hist}
    \centering
    \begin{tabular}{ccccc}
    \toprule
    Task & AD & DICP & DPT \\
    \midrule
     \textit{DarkRoom} & 14.1 & 11.4 & \textbf{19.6} \\
     \textit{Darkroom-Permuted} & \textbf{7.9} & 6.3 & -2.8 \\
     \textit{Darkroom-Large} & \textbf{22.3} & 0.9 & 16.4 \\
     \textit{Dark Key-to-Door} & 0.3 & \textbf{1.0} & 0.1 \\
    \midrule
    Average & \textbf{11.2} & 4.9 & 8.3 \\
    \bottomrule
    \end{tabular}
\end{table}

\subsection{ICRL with lightweight models}
\label{append_addition_result_lightweight_model}

\begin{figure}[ht]
\centering 
\setcounter{subfigure}{0}
\subfigure[Darkroom]
{
	\begin{minipage}{0.25\linewidth}
	\centering 
	\includegraphics[width=1.0\columnwidth]{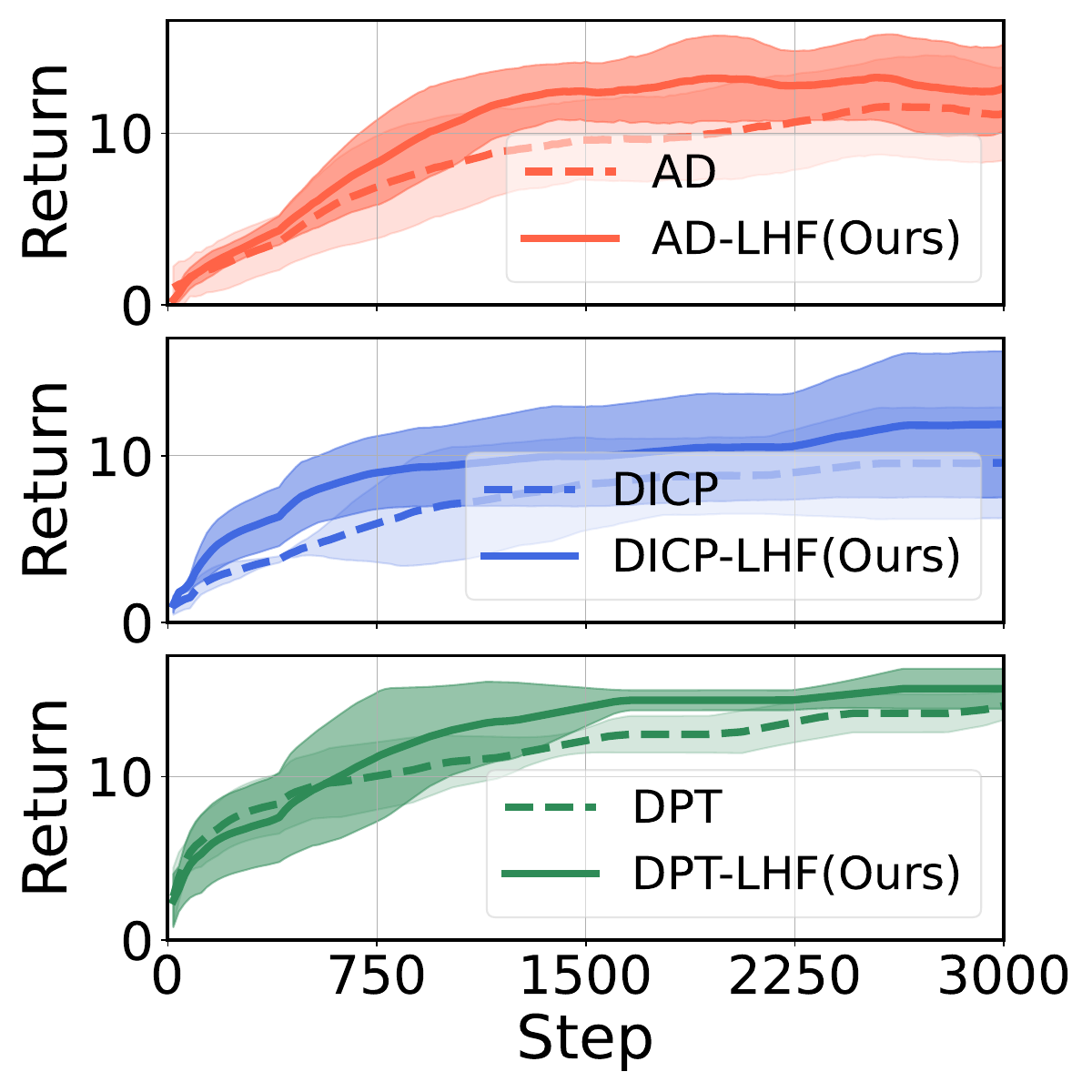}
	\end{minipage}
}
\!\!\!\!\!\!\!
\subfigure[Darkroom-Permuted]
{
	\begin{minipage}{0.25\linewidth}
	\centering 
	\includegraphics[width=1.0\columnwidth]{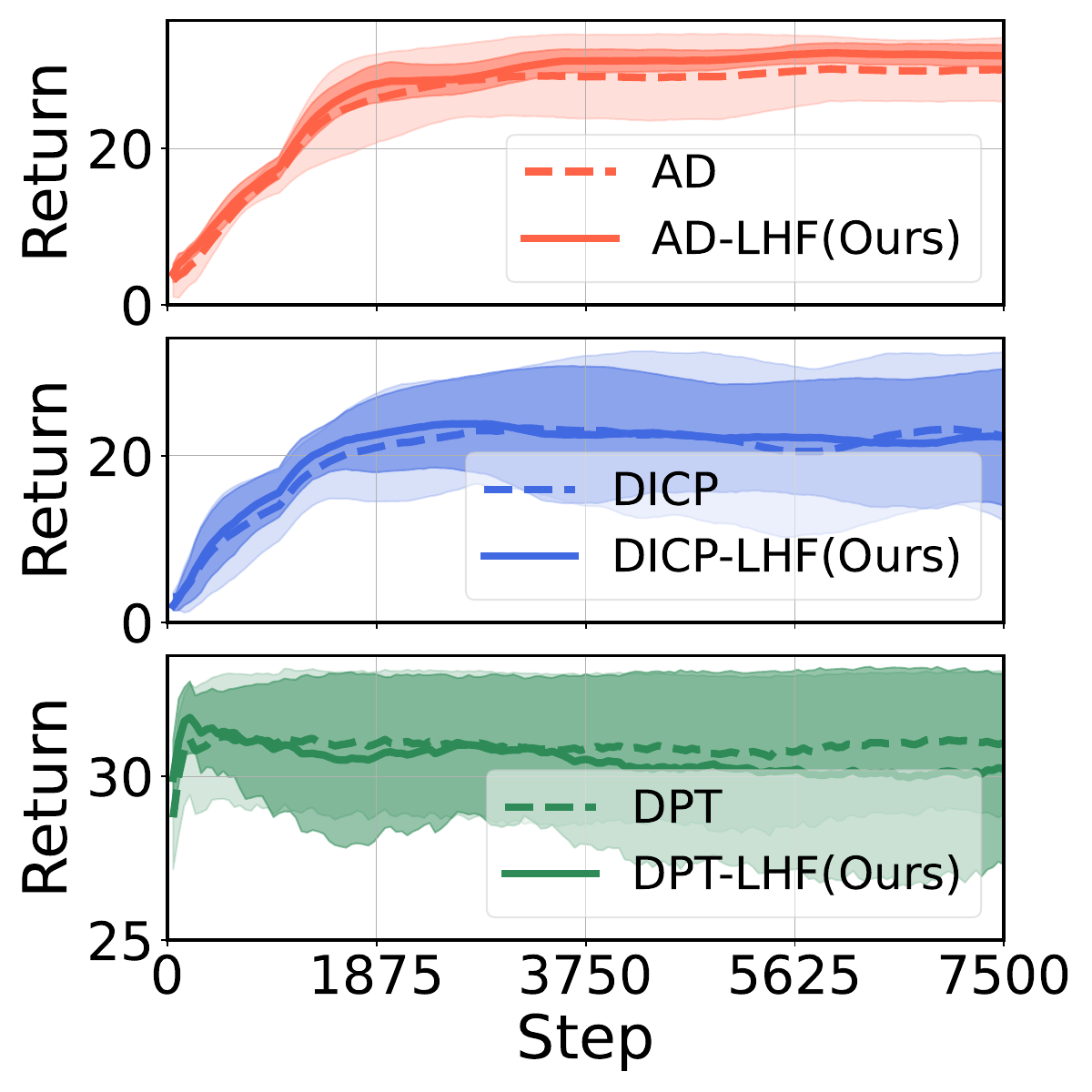}
	\end{minipage}
}
\!\!\!\!\!\!\!
\subfigure[Darkroom-Large]
{
	\begin{minipage}{0.25\linewidth}
	\centering 
	\includegraphics[width=1.0\columnwidth]{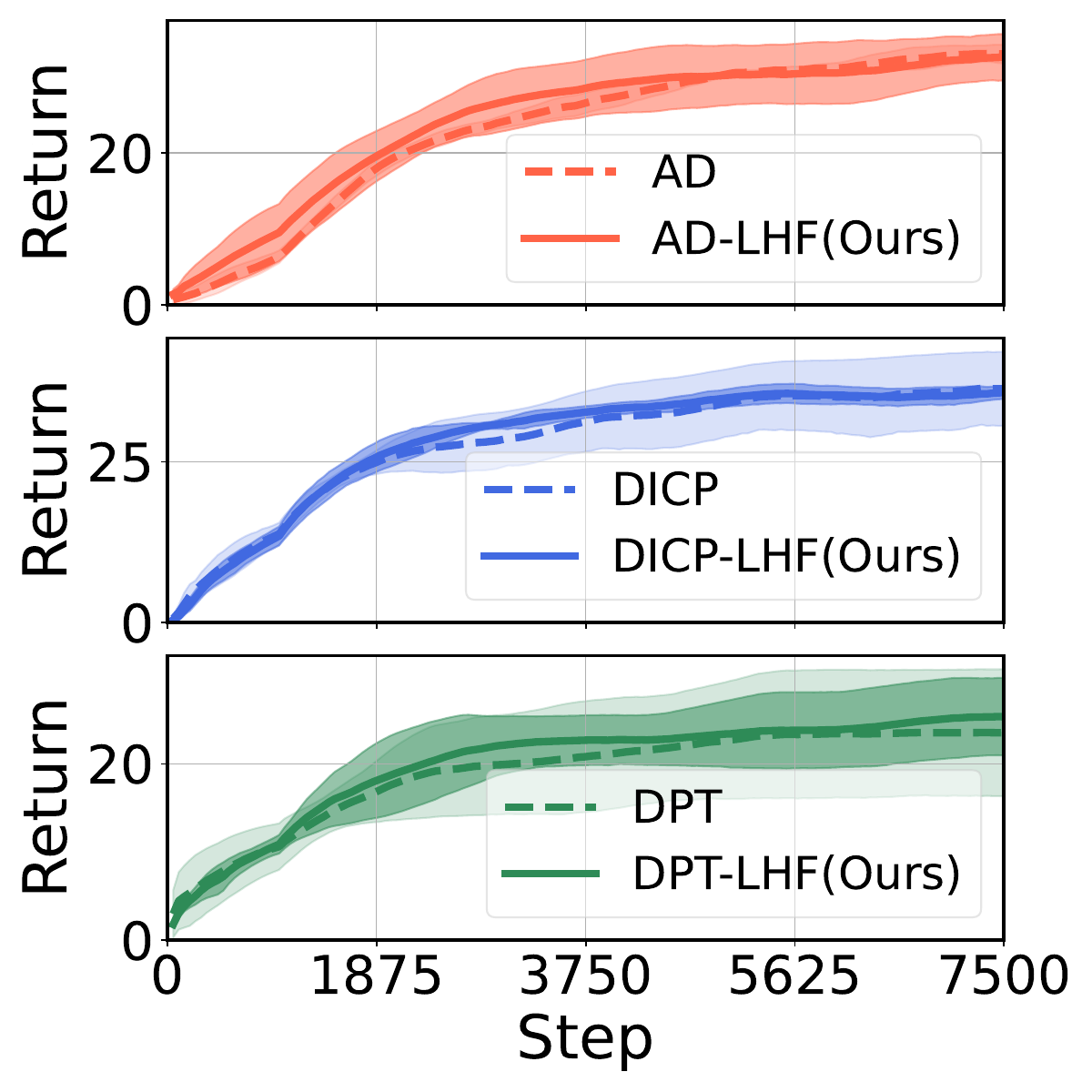}
	\end{minipage}
}
\!\!\!\!\!\!\!
\subfigure[Dark Key-to-Door]
{
	\begin{minipage}{0.25\linewidth}
	\centering 
	\includegraphics[width=1.0\columnwidth]{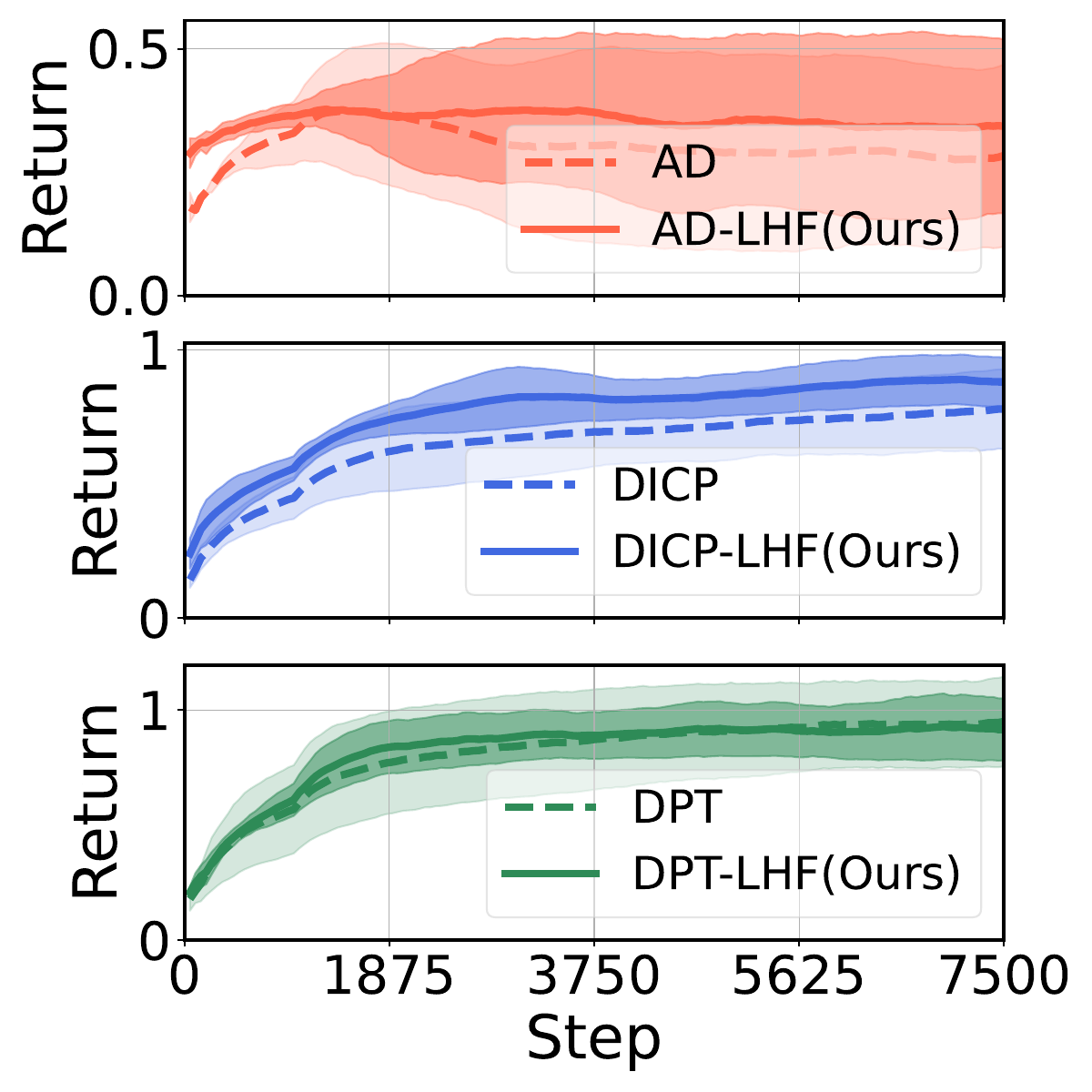}
	\end{minipage}
}
\caption{Learning curves of our LHF approach (solid lines) compared with original baselines (dashed lines) during the test. Each algorithm contains three independent runs with mean and standard deviation, provided with lightweight models. The backbone algorithms include AD (red), DICP (blue), and DPT (green).}
\label{fig_darkrooms_lightweight_model}
\end{figure}

Section~\ref{exp_numerical_result} also discusses the results of our LHF approach combined with a lightweight transformer model, which still demonstrates the superiority of LHF under the restricted model capacity. The detailed numerical results are referred to Table~\ref{tab_darkrooms_lightweight_model} and
Figure~\ref{fig_darkrooms_lightweight_model}.


\begin{table}[ht]
    \caption{Relative enhancement (\%) of our LHF approach over the baselines, provided with lightweight models. Backbone algorithms: AD, DICP, DPT.}
    \label{tab_darkrooms_lightweight_model}
    \centering
    \begin{tabular}{ccccc}
    \toprule
    Task & AD & DICP & DPT \\
    \midrule
     \textit{DarkRoom} & 22.4 & \textbf{28.8} & 10.6 \\
     \textit{Darkroom-Permuted} & \textbf{6.3} & 2.5 & -1.3 \\
     \textit{Darkroom-Large} & 4.9 & 2.1 & \textbf{5.4} \\
     \textit{Dark Key-to-Door} & 16.7 & \textbf{18.7} & 2.1 \\
    \midrule
    Average & 12.6 & \textbf{13.0} & 4.2 \\
    \bottomrule
    \end{tabular}
\end{table}

\subsection{Sensitivity analysis with respect to source RL algorithm}
\label{append_addition_result_sensitivity_sourceRL} 

\begin{figure}[ht]
\centering 
\setcounter{subfigure}{0}
\subfigure[Reach]
{
	\begin{minipage}{0.25\linewidth}
	\centering 
	\includegraphics[width=1.0\columnwidth]{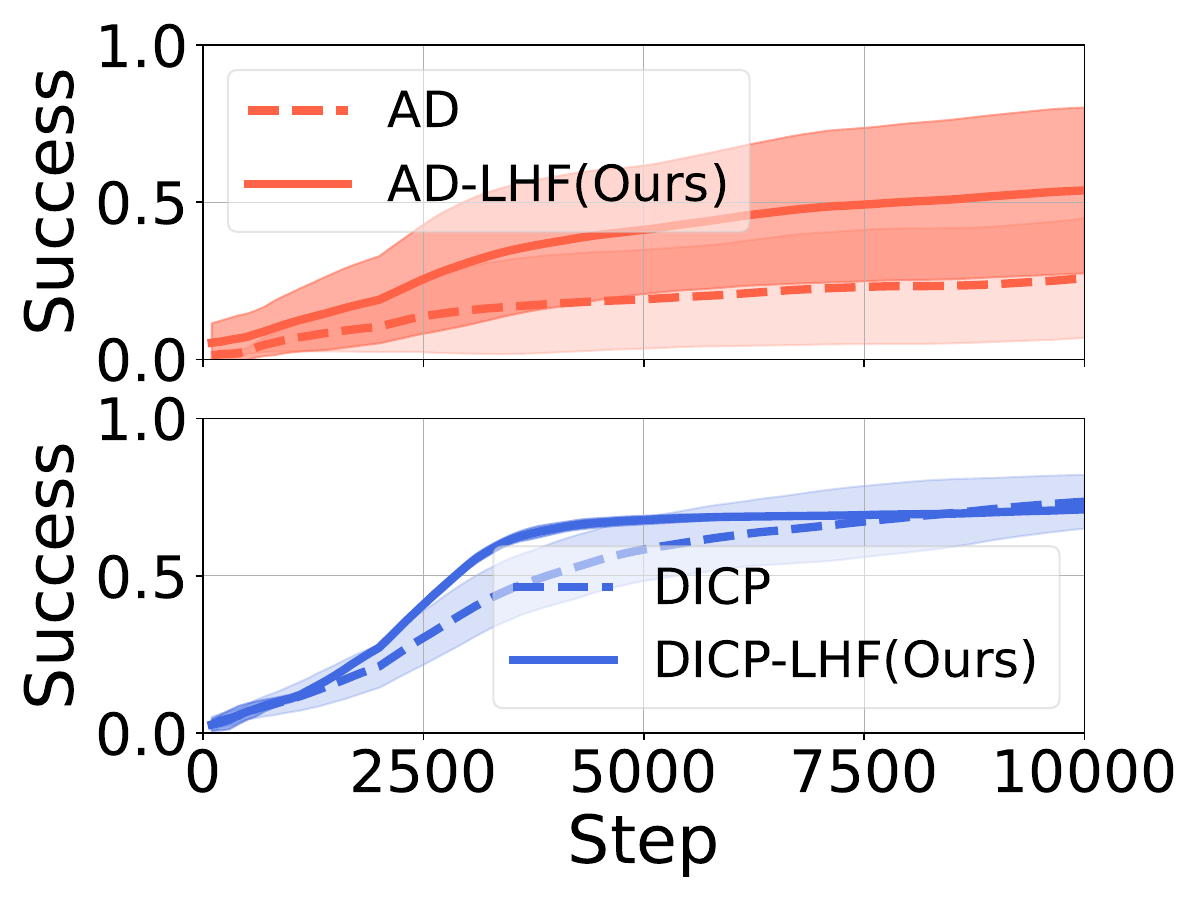}
	\end{minipage}
}
\!\!\!\!\!\!\!
\subfigure[Button-Press]
{
	\begin{minipage}{0.25\linewidth}
	\centering 
	\includegraphics[width=1.0\columnwidth]{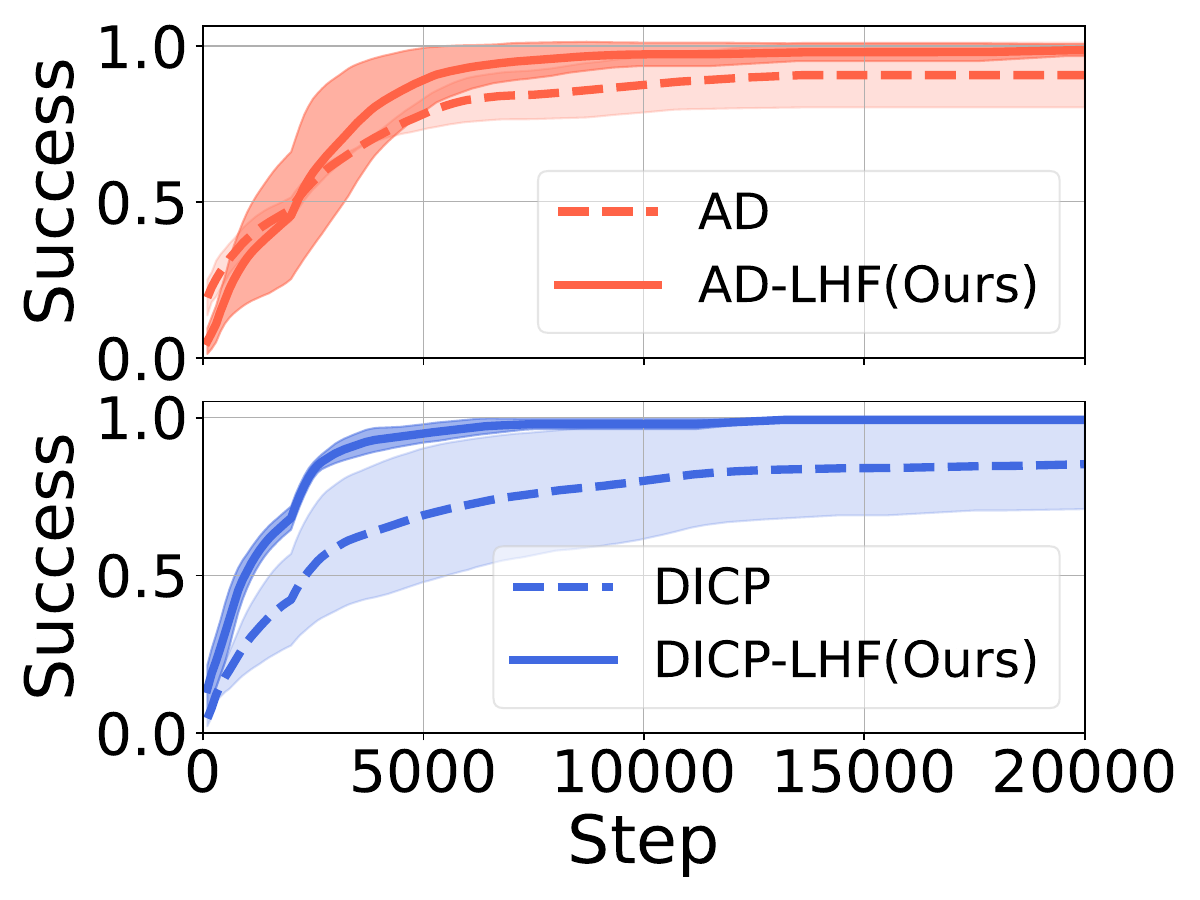}
	\end{minipage}
}
\!\!\!\!\!\!\!
\subfigure[Push]
{
	\begin{minipage}{0.25\linewidth}
	\centering 
	\includegraphics[width=1.0\columnwidth]{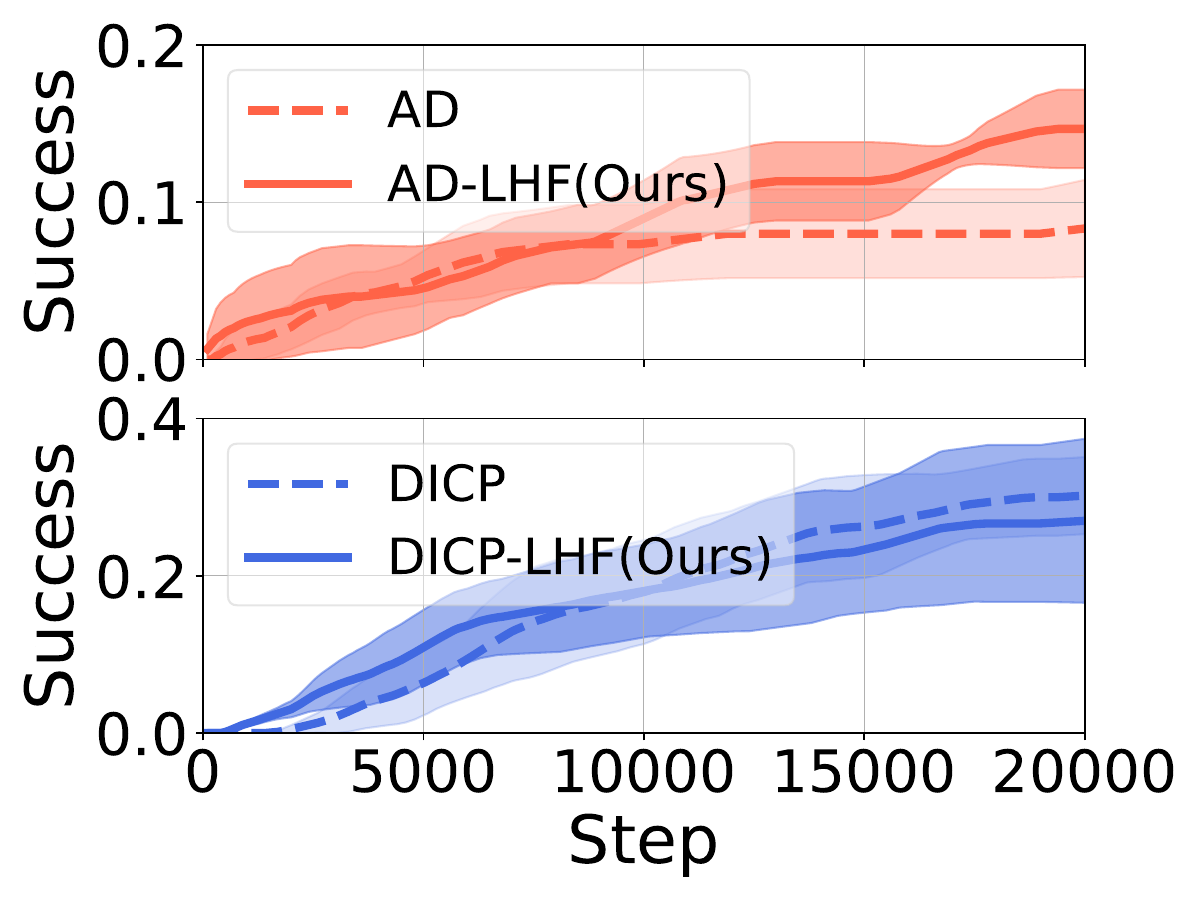}
	\end{minipage}
}
\!\!\!\!\!\!\!
\subfigure[Soccer]
{
	\begin{minipage}{0.25\linewidth}
	\centering 
	\includegraphics[width=1.0\columnwidth]{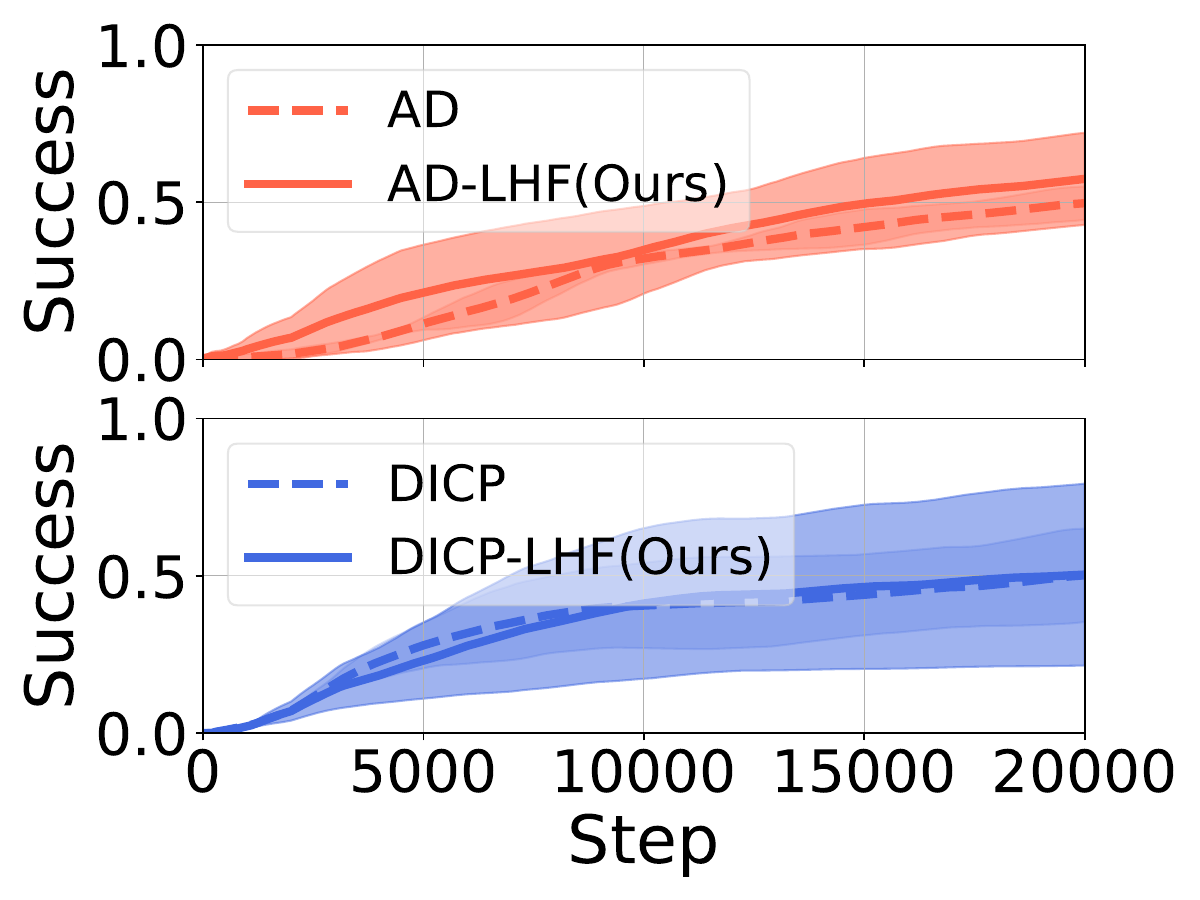}
	\end{minipage}
}
\caption{Learning curves of our LHF approach (solid lines) compared with original baselines (dashed lines) during the test. Each algorithm contains three independent runs with mean and std., provided with \textit{Meta-World-ML1} environments and datasets collected by SAC. The backbone algorithms include AD (red) and DICP (blue).}
\label{fig_sac_source_rl}
\end{figure}

Section~\ref{section_sensitivity} validates the robustness of our LHF performance in terms of the varying stability coefficient $\lambda$, distinct sampling strategies, and different source RL algorithms.
We present the numerical results of the first two in the main texts and exhibit the last one in Table~\ref{tab_sac_source_rl} and Figure~\ref{fig_sac_source_rl}.


\begin{table}[ht]
    \caption{Relative enhancement (\%) of our LHF approach over the baselines, provided with datasets collected by SAC. Backbone algorithms: AD and DICP.}
    \label{tab_sac_source_rl}
    \centering
    \begin{tabular}{cccc}
    \toprule
    Task & AD & DICP \\
    \midrule
     \textit{Reach} & \textbf{110.4} & 10.7 \\
     \textit{Button-Press} & 8.6 & \textbf{26.8} \\
     \textit{Push} & \textbf{33.5} & 0.0 \\
     \textit{Soccer} & \textbf{23.3} & -0.6 \\
     \midrule
     Average & \textbf{44.0} & 9.2 \\
    \bottomrule
    \end{tabular}
\end{table}

\section{Computing Infrastructure}
\label{Computing_Infrastructure}
All numerical experiments were conducted on a
workstation with
Intel\textregistered\ Core\texttrademark\ i9-14900KF CPU (32 threads),
and NVIDIA GeForce RTX 4090 GPU (24 GB), 64 GB RAM.

\section{Code}

The codes will be made available upon the publication of this work.

\end{document}